\documentclass[lettersize,journal]{IEEEtran}
\usepackage{amsmath,amsfonts}
\usepackage{algorithmic}
\usepackage{algorithm}
\usepackage{array}
\usepackage[caption=false,font=normalsize,labelfont=sf,textfont=sf]{subfig}
\usepackage{textcomp}
\usepackage{stfloats}
\usepackage{url}
\usepackage{verbatim}
\usepackage{graphicx}
\usepackage{cite}
\usepackage{amssymb}

\usepackage{booktabs}       %
\usepackage{nicefrac}       %
\usepackage{colortbl}
\usepackage{makecell}
\usepackage{multirow}
\usepackage{float}
\usepackage{pifont}
\usepackage{arydshln} %

\definecolor{Green}{rgb}{0.0,1.,0.0}
\definecolor{SpringGreen}{rgb}{0.1,0.9,0.0}
\definecolor{Yellow}{rgb}{0.5,0.5,0.0}
\definecolor{cvprblue}{rgb}{0.21,0.49,0.74}

\definecolor{tabfirst}{rgb}{1, 0.75, 0.7}
\definecolor{tabsecond}{rgb}{1, 0.83, 0.7}
\definecolor{tabthird}{rgb}{1, 0.96, 0.7}

\begin{document}

\newcommand{\orange}[1]{\textcolor{orange}{#1}}
\newcommand{\green}[1]{\textcolor{green}{#1}}
\newcommand{\blue}[1]{\textcolor{blue}{#1}}
\newcommand{\gray}[1]{\textcolor{gray}{#1}}
\newcommand{\fs}{\cellcolor{colorFst}\bf}   %
\newcommand{\nd}{\cellcolor{colorSnd}}      %
\newcommand{\rd}{\cellcolor{colorTrd}}      %
\newcommand{\lo}{\color{colorLow}}

\title{CAD-SLAM: Consistency-Aware Dynamic SLAM with Dynamic-Static Decoupled Mapping}


\author{Wenhua Wu, Chenpeng Su, Siting Zhu, Tianchen Deng, Jianhao Jiao, Guangming Wang, \\
Dimitrios Kanoulas, Zhe Liu, Hesheng Wang*~\IEEEmembership{Senior Member,~IEEE,}

\thanks{Wenhua Wu, Chenpeng Su, Siting Zhu, Tianchen Deng, and Zhe Liu are with the Department
 of Automation, Shanghai Jiao Tong University, China. Jianhao Jiao and Dimitrios Kanoulas are with the Department of Computer Science, University College London, UK. Guangming Wang is with the Department of Engineering, University of Cambridge, UK. Hesheng Wang is with the Department of Automation, Key Laboratory of System Control and Information
 Processing of Ministry of Education, State Key Laboratory of Avionics
 Integration and Aviation System-of-Systems Synthesis, Shanghai Key Laboratory of Navigation and Location Based Services, Shanghai Jiao Tong
 University, Shanghai, China.}
 \thanks{* Corresponding Author}

 }



\maketitle

\begin{abstract}
Recent advances in neural radiation fields (NeRF) and 3D Gaussian-based SLAM have achieved impressive localization accuracy and high-quality dense mapping in static scenes. However, these methods remain challenged in dynamic environments, where moving objects violate the static-world assumption and introduce inconsistent observations that degrade both camera tracking and map reconstruction. This motivates two fundamental problems: robustly identifying dynamic objects and modeling them online. 
To address these limitations, we propose CAD-SLAM, a \underline{C}onsistency-\underline{A}ware \underline{D}ynamic SLAM framework with dynamic-static decoupled mapping. Our key insight is that dynamic objects inherently violate cross-view and cross-time scene consistency. We detect object motion by analyzing geometric and texture discrepancies between historical map renderings and real-world observations. Once a moving object is identified, we perform bidirectional dynamic object tracking (both backward and forward in time) to achieve complete sequence‑wise dynamic recognition. Our consistency-aware dynamic detection model achieves category-agnostic, instantaneous dynamic identification, which effectively mitigates motion-induced interference during localization and mapping. In addition, we introduce a dynamic–static decoupled mapping strategy that employs a temporal Gaussian model for online incremental dynamic modeling. Experiments conducted on multiple dynamic datasets demonstrate the flexible and accurate dynamic segmentation capabilities of our method, along with the state-of-the-art performance in both localization and mapping.

\end{abstract}


\section{Introduction}
\label{sec:intro}

\subsection{Motivations}

\IEEEPARstart{D}{ense} visual Simultaneous Localization and Mapping (SLAM) is the foundation of perception, navigation, and planning that finds wide applications in areas such as autonomous driving, mobile robotics, and virtual reality~\cite{wang2024survey, al2024review}. SLAM consists of two primary components: estimating the position of the sensing system within an unknown environment, and constructing a map of that environment.

Although existing dense visual SLAM methods based on Neural Radiance Field (NeRF) and 3D Gaussians have shown promising results in static scenes, their performance often deteriorates in complex dynamic environments, especially when there are highly dynamic objects in the observations~\cite{tosi2024nerfs}. This degradation arises primarily because such highly-dynamic objects severely violate the static-world assumption underlying both geometric and photometric optimization. Their rapid and irregular motion produces prominent inconsistent inter-frame correspondences, which directly corrupts camera pose estimation and distorts the recovered 3D geometry~\cite{carlone2025slam}.

\subsection{Challenges}

To support robots in maintaining stability within dynamic environments, the core objective of dynamic SLAM is not only to estimate camera poses and reconstruct static backgrounds but also to support the reliable perception and modeling of dynamic elements, allowing downstream tasks such as safe navigation, obstacle avoidance, and human-robot interaction~\cite{carlone2025slam}. This objective presents two inherent challenges.

The first challenge is \textit{correctly and completely identifying general dynamic objects}. Real-world environments exhibit open-world semantic characteristics. dynamic objects are not limited to predefined categories (e.g. pedestrians, vehicles) but also include arbitrary moving entities (e.g., manipulated boxes, floating balloons). This renders semantic-prior-based methods~\cite{schischka2024dynamon, xu2024nid, ruan2023dn, li2025pg} ineffective for unseen categories and incapable of accurately distinguishing motion states, for instance, differentiating between a parked car and a moving car. Furthermore, observation changes induced by camera movement are easily confounded with the actual motion of objects. As a result, methods utilizing optical flow~\cite{jiang2024rodyn} or multi-view depth warping masks~\cite{xu2024dg} fail to accurately segment truly dynamic objects. Third, ambiguity of the motion boundary. fast-moving objects produce blurred boundaries in consecutive frames, and low-texture regions further exacerbate the difficulty of precise delineation~\cite{zheng2025wildgs}. These limitations highlight the need for a dynamic object identification mechanism that is independent of brittle semantic priors while capable of capturing fine-grained motion boundaries.

The second challenge is \textit{effectively tracking and modeling dynamic objects}. Dynamic objects are essential for understanding the environment in downstream interactions. However, dynamic modeling faces several obstacles. First, the coupling between the motion of the camera and the object. The SLAM system must simultaneously estimate the camera pose and model object motion, where errors in either task propagate to the other. Second, real‑time performance constraints. online SLAM requires that dynamic modeling be completed within the frame interval, which precludes heavy offline optimization methods~\cite{luiten2024dynamic, wu20244d, yangreal} that rely on pre‑computed poses or full‑sequence data. Finally, moving objects may undergo deformation, which requires the modeling framework to adapt to temporal changes. These fundamental difficulties highlight the urgent need for an online, SLAM‑native dynamic modeling solution—one that can decouple camera and object motion while efficiently capturing temporal dynamics.

\subsection{Contributions}

The aforementioned challenges still remain unsolved, directly motivating our CAD-SLAM, an adaptive dynamic dense visual SLAM with Gaussian splatting. Our key \textbf{insight} is that dynamic objects inherently introduce inconsistency between historical mapping and actual observation. To be specific,  observations from different viewpoints and timestamps exhibit consistency under static-world assumption. The movement of objects breaks this consistency, leading to discrepancies in both geometry and texture between the rendered historical map and the real-time observation. We then leverage this clue to prompt fine-grained and class-agnostic segmentation. Unlike semantic-based strategies or iterative dynamic identification pipelines, our dynamic identification mechanism enables both high adaptiveness and instant responsiveness.

\begin{figure*}[t] 
\center{
\includegraphics[width=\linewidth]{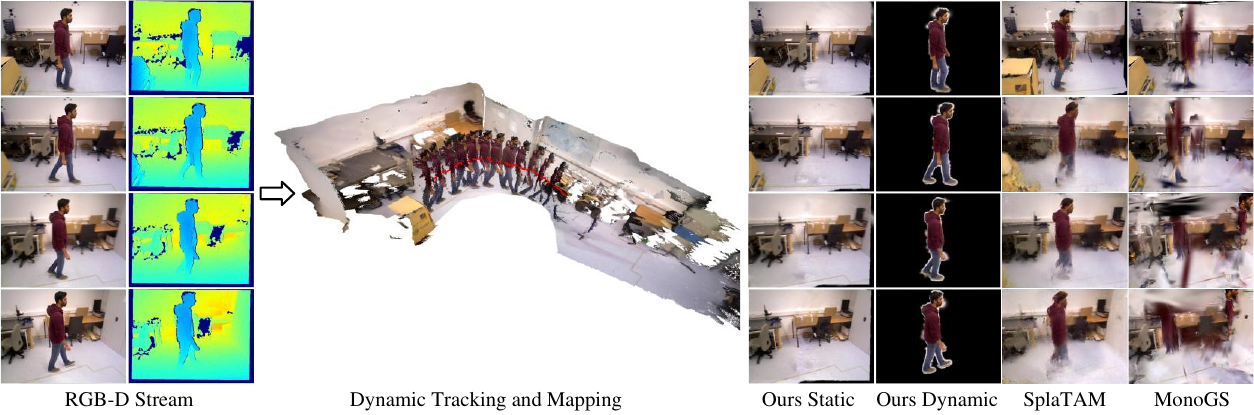}}
        \caption{ 
         CAD-SLAM. Given RGB-D stream, our method achieves precise camera pose tracking while constructing dynamic-static decoupled maps. Our method can adaptively segment dynamic objects of any category without any semantic priors. The illustration presents effective tracking and mapping performance under dynamic scenarios.
        }
    \label{fig:teaser}
\end{figure*}

Upon detecting a dynamic object, we initiate a bidirectional tracklet for concise and effective tracking, endowing the SLAM system with precise and timely dynamic cues. In camera tracking, dynamic object pixels are excluded to mitigate motion-induced corruption, ensuring more accurate and robust pose estimation. Beyond mere suppression of dynamic regions, we propose a dynamic–static composite mapping strategy. For dynamic objects, we construct a temporal Gaussian model that supports online incremental modeling, capturing their motion and appearance evolution over time. For the static background, dynamic regions are removed, and previously occluded areas are progressively completed as the object moves, enabling the static map to grow cleanly and consistently over time.


Overall, the main contributions of this paper are as follows:
\begin{itemize}
\item We propose CAD-SLAM, a novel dynamic dense visual SLAM that accurately performs tracking and mapping in complex dynamic environments while simultaneously tracking and modeling dynamic objects.

\item We introduce a semantic-free, consistency-aware dynamic identification method that enables category-agnostic, fine-grained, and instantaneous detection of dynamic objects without any priors or iterative learning.

\item We develop a hybrid static–dynamic mapping framework that incorporates a temporal Gaussian model for online incremental modeling of dynamic objects while maintaining a clean and progressively completed static background.

\item Extensive experiments conducted on multiple real-world dynamic datasets demonstrate that our method achieves state-of-the-art performance in both camera tracking and dense mapping, while exhibiting strong adaptability across diverse scenarios.
\end{itemize}


\section{Related Work}
\label{sec:related_work}

\subsection{Traditional Visual SLAM} 
The foundation of dense visual SLAM lies in DTAM~\cite{newcombe2011dtam}, which pioneers real-time tracking and mapping via dense scene representation. In the same period, the KinectFusion method~\cite{newcombe2011kinectfusion} makes notable strides by using ICP algorithms~\cite{besl1992method} and volumetric TSDF to achieve accurate and real-time reconstruction of dense surfaces for indoor scenes. Several data structures, including Surfels~\cite{whelan2015elasticfusion,schops2019surfelmeshing} and Octrees~\cite{vespa2018efficient,xu2019mid}, have been proposed to improve scalability and reduce memory. In contrast to these methods, which rely on per-frame pose optimization, BAD-SLAM~\cite{schops2019bad} is the first to propose a full Bundle Adjustment (BA) to jointly optimize the keyframes. 
Traditional visual SLAM has matured over decades into a well‑established framework, with representative systems such as ORB‑SLAM~\cite{mur2015orb,mur2017orb, campos2021orb}, DSO~\cite{engel2017direct}, and VINS‑Mono~\cite{qin2018vins}. These systems typically follow a pipeline of feature extraction and matching, pose estimation, local mapping, and loop closure detection, relying on sparse feature points (e.g. ORB~\cite{rublee2011orb}, SIFT~\cite{lowe2004distinctive}) or direct pixel intensities to optimize camera poses and sparse 3D structure. 
In recent years, numerous deep learning-based SLAM methods~\cite{bloesch2018codeslam, li2020deepslam, koestler2022tandem, peng2020convolutional, teed2021droid} have been introduced to improve the precision and robustness of traditional SLAM methods. However, the point or mesh maps generated by traditional SLAM do not meet the demands of high-density reconstruction and photorealistic rendering. 



\subsection{Neural Implicit SLAM}
The advent of Neural Radiance Fields (NeRF)~\cite{mildenhall2021nerf} has revolutionized scene representation by modeling scenes as implicit neural networks, enabling high-fidelity dense reconstruction. This paradigm naturally overcomes the problem of the sparse structure of traditional SLAM, spurring the development of NeRF-based visual SLAM. iMAP~\cite{sucar2021imap} pioneered the integration of NeRF into the SLAM framework, achieving joint optimization of camera poses and an implicit map. However, its single Multi-Layer Perceptron (MLP) design often restricts reconstruction detail and is prone to catastrophic forgetting. This challenge inspired NICE‑SLAM~\cite{zhu2022nice} to introduce a hierarchical feature‑grid scene representation, improving scalability and efficiency. Co‑SLAM~\cite{wang2023co} combines the smoothness and fast convergence of coordinate encoding with the local‑detail representation advantages of sparse parametric encoding. ESLAM~\cite{johari2023eslam} improves SLAM performance by implementing multi-scale axis-aligned feature planes and using TSDF for faster and more refined mapping. Subsequent studies~\cite{ yang2022vox, Zhang_2023_ICCV, liu2023multi, li2023end, ming2022idf, wang2024structerf, wu2024dvn, deng2024plgslam} have further advanced scene representation and camera tracking through a series of refinements.
However, NeRF-based SLAM still faces a critical bottleneck: the computational overhead of ray sampling and MLP inference is substantial, making it difficult to meet the demands of real-time applications.

 \subsection{3D Gaussian Splatting SLAM}
 3D Gaussian Splatting (3DGS)~\cite{kerbl20233d} is a breakthrough scene representation that has emerged in recent years, combining the explicit nature of traditional point clouds with the high-quality rendering capability of NeRF~\cite{deng2025best3dscenerepresentation}. It represents a scene as a set of 3D Gaussians, each parameterized by its position, covariance, and radiance attributes, and achieves real-time photorealistic rendering through a splatting operation. This effectively addresses the efficiency bottleneck of NeRF. Due to these advantages, 3DGS has been rapidly integrated into the field of visual SLAM, leading to a series of 3DGS-based visual SLAM methods~\cite{deng2025vpgs}.
 GS-SLAM~\cite{yan2024gs} was the first to employ fast splatting techniques and introduced a dynamic, adaptive 3D Gaussian expansion strategy. Photo-SLAM~\cite{huang2024photo} integrates both explicit geometric features and implicit texture representations, utilizing geometry-based densification and Gaussian-pyramid-based learning within a multi-threaded framework. Gaussian-SLAM~\cite{yugay2023gaussian} organizes scenes into independently optimized sub-maps, enhancing efficiency and scalability. SplaTAM~\cite{keetha2024splatam} and GSSLAM~\cite{Matsuki:Murai:etal:CVPR2024} represent scenes using 3DGS and fundamentally redefine dense SLAM processes, further enhancing robustness by incorporating geometric verification, regularization techniques, and covisibility-based keyframe selection. Although these methods have achieved impressive performance in static scenes, they struggle in dynamic environments primarily because dynamic objects violate scene consistency.

\subsection{SLAM in Dynamic Environments.}
The interference of dynamic objects has long been one of the core challenges in the development of SLAM technology. For traditional SLAM, on the one hand, object motion leads to erroneous feature matches, causing pose drift; on the other hand, dynamic regions introduce outliers into bundle adjustment, degrading optimization accuracy. Existing improvements mostly rely on semantic segmentation or optical flow to identify dynamic areas, which are then removed during feature matching and bundle adjustment~\cite{bescos2018dynaslam, yu2018ds, bescos2021dynaslam, zhang2020vdo,xiao2019dynamic, zhang2020flowfusion, deng2025mne}.  Similarly, for neural implicit SLAM and 3DGS SLAM, object motion induces photometric and geometric inconsistencies between frames, resulting in localization drift and rendering artifacts.

To handle dynamics, several methods~\cite{ruan2023dn, li2024ddn, xu2024nid} employ semantic segmentation or object detection. However, they rely on predefined dynamic category priors, leading to two inherent limitations: (1) inability to process unknown object categories, and (2) misclassification of stationary objects from predefined dynamic categories as moving objects. DynaMoN~\cite{schischka2024dynamon} and RoDyn-SLAM~\cite{jiang2024rodyn} incorporate optical flow estimation, but the inherent ambiguity between object motion and camera motion measurements remains challenging to disambiguate. DG-SLAM \cite{xu2024dg} designs a multi-view depth warp mask to compensate for missing objects. However, the occlusion caused by the view change is contained in the mask.
Gassidy \cite{wen2024gassidy} performs instance segmentation of the scene and relies on object-by-object iterative analysis to distinguish dynamics. The latest WildGS-SLAM \cite{zheng2025wildgs} introduces an uncertainty-aware approach that eliminates dependency on prior. However, uncertainties in dynamic object boundaries remain prone to ambiguity, and the incrementally trained MLP performs poorly at the beginning stage and short sequence case. In contrast, our method adaptively segments arbitrary dynamic objects through scene consistency analysis, requiring no prior knowledge while achieving precise boundary delineation. Furthermore, while existing methods simply filter out dynamic elements to construct static maps only, ours simultaneously builds both dynamic and static maps.

\section{Method}\label{sec:method}
\begin{figure*}[t] 
\center{\includegraphics[width=1.0\textwidth]{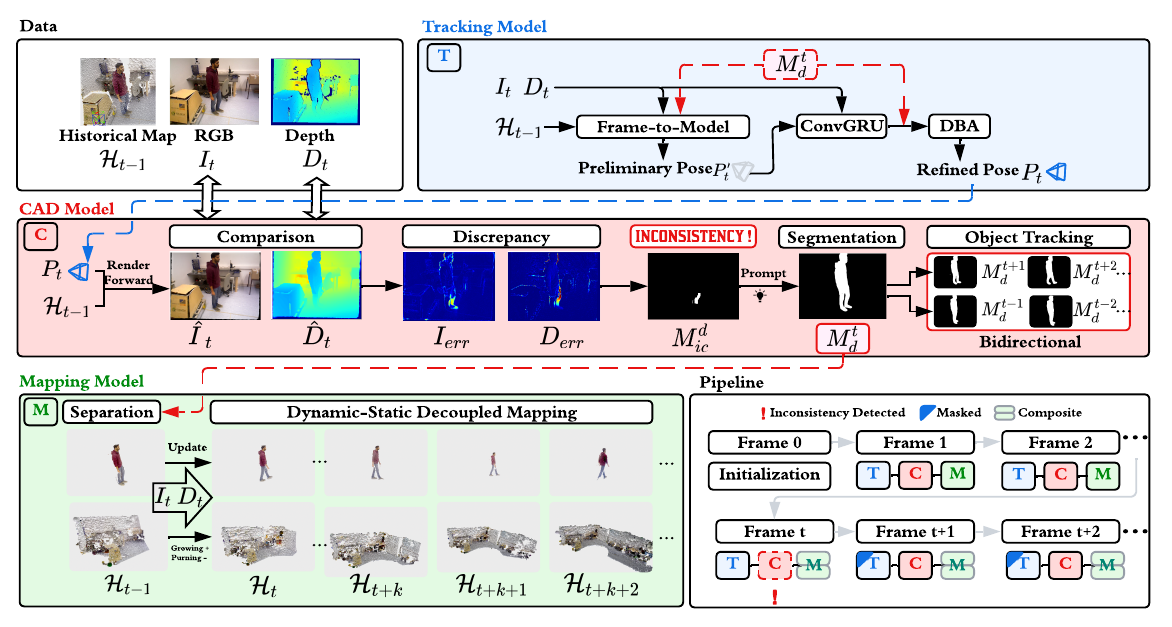}} 
\caption{Overview of our complete SLAM system. CAD-SLAM consists of three tightly coupled models and takes RGB-D as input to produce camera poses and a dense Gaussian map. For each incoming frame, the tracking model first predicts an initial pose using frame-to-model tracking, then refines it with a pretrained ConvGRU after DBA. Using this refined pose, the Consistency-Aware Dynamic (CAD) model renders the historical map forward and compares it with the current observation to compute an inconsistency map that reveals dynamic regions. Guided by this prompt, CAD model adaptively segments dynamic masks and performs bidirectional temporal tracking of dynamic objects. The dynamic masks allow the mapping model to separate static and dynamic components and maintain dynamic–static decoupled mapping. Throughout SLAM, whenever CAD model detects strong inconsistency and produces or updates the mask, both tracking and mapping are subsequently guided by it.}
\label{framework}
\end{figure*}

Our proposed adaptive dynamic dense SLAM framework is illustrated in Fig.~\ref{framework}. The input consists of an RGB-D image sequence $\{(\mathbf{I}_t, \mathbf{D}_t)|, t = 0,1,\dots,n\}$ and camera intrinsics $\mathbf{K}$, while the output includes estimated camera poses and a dynamic–static decoupled map. The system comprises three tightly coupled modules: a tracking module for pose estimation, a CAD module for dynamic object identification, and a mapping module for dynamic-static decoupled dense reconstruction. The pipeline begins with static map initialization using the first frame (Sect.~\ref{m1}). As SLAM progresses, dynamic objects manifest motion and are adaptively segmented via scene-consistency analysis, triggering bidirectional dynamic object tracking (Sect.~\ref{m2}). Based on accurate dynamic identification, dynamic–static separation is applied to the initial static map. Camera poses are subsequently optimized using the masked tracking loss (Sect.~\ref{m3}), followed by dynamic–static decoupled mapping to update both the static background and dynamic object models (Sect.~\ref{m4}).

\subsection{3D Gaussian Splatting}
\label{m1}
We adopt 3D Gaussian Splatting~\cite{kerbl20233d} as the fundamental scene representation for our dynamic SLAM system. The scene is represented as a set of 3D Gaussians $\mathcal{G} = \{G_i\}_{i=1}^N$, where each Gaussian $G_i$ is parameterized by its mean position $\boldsymbol{\mu}_i \in \mathbb{R}^3$, covariance matrix $\boldsymbol{\Sigma}_i \in \mathbb{R}^{3 \times 3}$, opacity $o_i \in [0,1]$, and spherical harmonic (SH) coefficients $\boldsymbol{h}_i \in \mathbb{R}^{k}$ for view-dependent color modeling. The spatial influence of the $i$-th Gaussian is defined by the probability density function:
\begin{equation}
    g_i(\mathbf{x}) = \exp\left(-\frac{1}{2}(\mathbf{x}-\boldsymbol{\mu}_i)^\top \boldsymbol{\Sigma}_i^{-1}(\mathbf{x}-\boldsymbol{\mu}_i)\right).
\end{equation}
Following standard practice, the covariance matrix $\boldsymbol{\Sigma}_i$ is decomposed into a rotation matrix $\mathbf{R}_i$ and a scaling matrix $\mathbf{S}_i$ to ensure semi-positive definiteness during optimization, such that $\boldsymbol{\Sigma}_i = \mathbf{R}_i \mathbf{S}_i \mathbf{S}_i^\top \mathbf{R}_i^\top$. Following prior Gaussian-based SLAM approaches, we initialize the static Gaussian map from the first RGB-D frame. Given the depth image and camera intrinsics, the initial point cloud is reconstructed and each point is converted into a 3D Gaussian.

\noindent\textbf{Differentiable Rendering.}
To render the scene from a camera pose $\mathbf{T}$, 3D Gaussians are projected onto the 2D image plane. The 2D covariance matrix $\boldsymbol{\Sigma}'_i$ in the screen space is approximated using the affine transformation of the view transformation:
\begin{equation}
    \boldsymbol{\Sigma}'_i = \mathbf{J} \mathbf{W} \boldsymbol{\Sigma}_i \mathbf{W}^\top \mathbf{J}^\top,
\end{equation}
where $\mathbf{J}$ is the Jacobian of the affine approximation of the projective transformation, and $\mathbf{W}$ is the viewing transformation matrix calculated from $\mathbf{T}$.

The rendered pixel values are computed using front-to-back alpha blending. Specifically, we sort the Gaussians based on their depth in the camera frame. The final color $\hat{\mathbf{C}}$, depth $\hat{D}$, and opacity mask $\hat{O}$ for a pixel are computed by accumulating the contributions of $N$ ordered Gaussians overlapping the pixel:
\begin{align}
    \hat{I} &= \sum_{i \in \mathcal{N}} c_i \alpha_i \prod_{j=1}^{i-1}(1-\alpha_j), \label{eq:render_color} \\
    \hat{D} &= \sum_{i \in \mathcal{N}} d_i \alpha_i \prod_{j=1}^{i-1}(1-\alpha_j), \label{eq:render_depth} \\
    \hat{O} &= \sum_{i \in \mathcal{N}} \alpha_i \prod_{j=1}^{i-1}(1-\alpha_j), \label{eq:render_opacity}
\end{align}
where $c_i$ is the view-dependent color computed from SH coefficients, $d_i$ is the projected depth of the Gaussian center, and $\alpha_i$ is the effective opacity derived from $o_i$ and the 2D Gaussian probability. This differentiable formulation allows for the simultaneous optimization of geometry and appearance via gradient descent.

\subsection{Consistency-Aware Dynamic Detection Model}
\label{m2}

\begin{figure*}[t] 
\center{\includegraphics[width=1.0\textwidth]{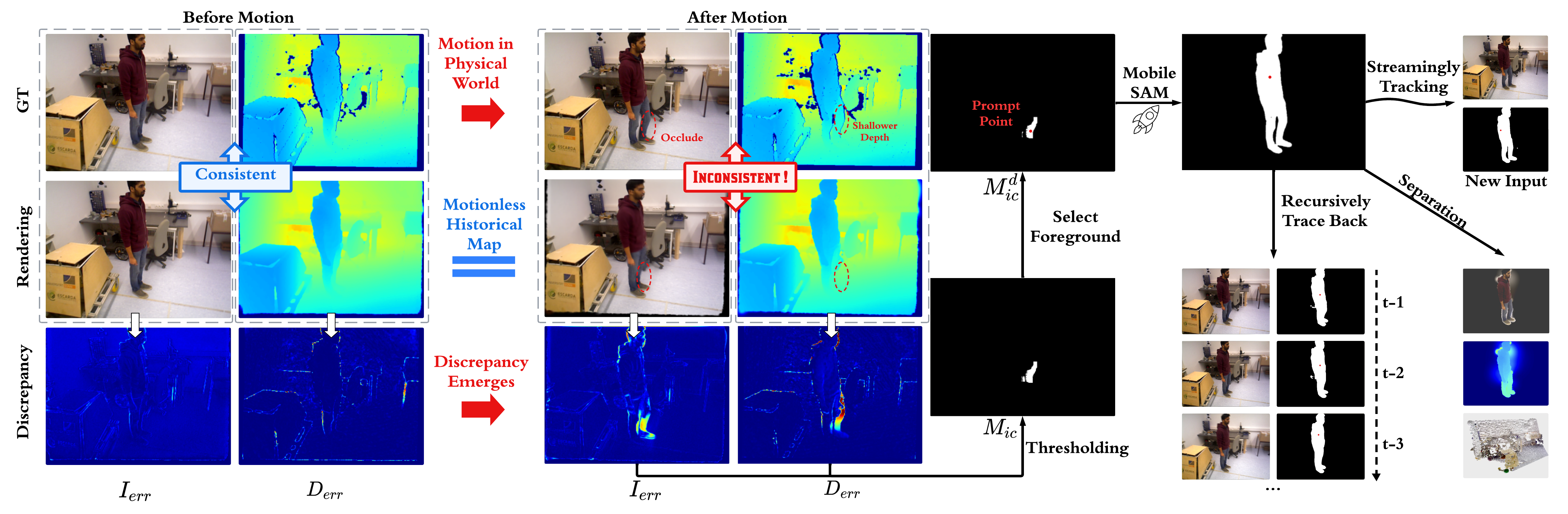}} 
\caption{CAD Model. Motion disrupts scene consistency, causing photometric and geometric discrepancies between the historical map and current observations. These inconsistencies are used to prompt MobileSAM for fine-grained mask extraction, facilitating bidirectional object tracking and dynamic–static separation.
}
\label{render-visw}
\end{figure*}

Most existing dynamic object recognition methods~\cite{li2025pg, xu2024dg, xu2024nid, ruan2023dn} rely heavily on pre-trained detection and semantic segmentation models. However, these approaches are inherently limited to a closed set of predefined dynamic categories, making SLAM systems weak in open-world scenarios. While recent frameworks attempt to mitigate this dependency by leveraging loss analysis or uncertainty learning, they often incur non-negligible computational overhead or require extra online training. To address these limitations, we propose a Consistency-Aware Dynamic (CAD) model. Crucially, our approach operates without semantic priors and maintains high computational efficiency, ensuring the low latency required for real-time SLAM.

\noindent \textbf{Scene Consistency Analysis.}
Assuming that the motion occurs at $t$, the historical Gaussian map $\mathcal{H}_{t-1}$ therefore represents a static reconstruction of the scene. Under the static assumption, the historical map accurately models the physical scene, which means that the renderings ($\hat{I}$,$\hat{D}$) of $\mathcal{H}_{t-1}$ should exhibit consistency with current observations ($I$,$D$). However, the presence of dynamic objects violates this alignment: the static map built from former $t-1$ frames fails to reflect the motion taking place at $t$. When rendered from pose estimated at $t$, significant discrepancies are produced in the areas undergoing motion. This inconsistency serves as a robust and generalizable cue for identifying dynamic regions within the scene. To capture the misalignment comprehensively, we simultaneously consider both photometric and geometric inconsistencies which are inherently complementary:
\begin{equation}
I_{err} = \lVert I - \hat{I}\rVert_2, \quad D_{err} = \lVert D -\hat{D}\rVert_1,
\end{equation}
\begin{equation}
M_{ic} = (I_{err} > \tau_I) \cup (D_{err} > \tau_D),
\end{equation}
where $\tau_I$ and $\tau_D$ represent the thresholds for color and geometric inconsistencies, respectively. $M_{ic}$ represents the mask of the inconsistent region.

\noindent \textbf{Dynamic Regions Detection.}
Dynamic objects in the field of view can be classified into two types: one is a dynamic object moving from outside the field of view into it, and the other is an object moving within the field of view. For the first type, the inconsistent region in the observed image corresponds to the dynamic object that has entered the field of view. For the second type, the inconsistent region in the observed image manifests as the area where the dynamic object has newly moved or the previously occluded background has been exposed. The new occlusion caused by the dynamic object in the background will make the observed depth smaller than the rendered depth. In contrast, the newly exposed background region will have an observed depth larger than the rendered depth. This allows for the distinction between dynamic objects and the background. 
\begin{equation}
M_{ic}^d = M_{ic} \cap ((D - \hat{D})<0),
\end{equation}
where $M_{ic}^d$ denotes the inconsistent region corresponding to dynamic objects.

\noindent \textbf{Dynamic Segmentation.} With the dynamic regions detected, we then utilize an off-the-shelf model to perform detailed segmentation. Considering efficiency and latency, we choose MobileSAM\cite{zhang2023faster}, a lightweight and fast segment model guided by prompts. Due to the minimal object movement between consecutive frames (with a time interval of 1/30s), the resulting dynamic inconsistency region represents only a small portion of the object. Therefore, we use the center of the inconsistent region as a prompt input for MobileSAM $f_{\theta}$ to obtain the complete dynamic object segmentation.
\begin{equation}
M_d = f_{\theta}(I,\textit{o}( M_{ic}^d)),
\label{eq:mobile_sam}
\end{equation}
where $\textit{o}(\cdot)$ represents the center of the inscribed circle.

\noindent \textbf{Dynamic Object Tracking.} 
For all detected dynamic objects $\mathcal{D}$, a unique ID is assigned to each instance, and we maintain an object state directory recording their historical states as the tuple $(\mathbf{C}_i, \mathbf{A}_i, \mathcal{M}_i)$. Specifically, $\mathcal{M}_i=\{M_i^k\}^t_{k=0}$ denotes the set of dynamic segmentation masks across time; $\mathbf{C}_i=\{c_i^k\}^t_{k=0}$ stores the sequence of mask centers; and $\mathbf{A}_i=\{a_i^k\}^t_{k=0}$ is the temporal visibility vector indicating the presence of object $i$ at each timestamp $k \in [0, t]$. For clarity, we denote the newly detected object at time $t$ by dropping the object index $i$ in the following analysis.

Since dynamic objects are detected via inconsistency analysis, the detection typically occurs after the object has initiated motion. To achieve a complete capture of the object's motion history, we employ a bidirectional tracking mechanism. For past frames($k<t$), we recursively trace back:
$M^{k-1}=f_\theta(I_{k-1},c_k)$ until the object exits the view frustum or the initial frame($k=0$). For future frames ($k>t$), the tracking process is integrated streamingly into the SLAM pipeline: $M_d^{t+1}=f_\theta(I_{t+1},c_t)$.



The bidirectional tracking mechanism not only reconstructs a complete temporal sequence of dynamic masks, but also provides reliable cross-frame associations for each object. Such temporally coherent trajectories are crucial for masked pose estimation and decoupled mapping.

In contrast to frame-wise semantic detection, which overlooks the temporal coherence of dynamic objects and remains constrained by closed-set category assumptions, our CAD module provides a more principled and efficient alternative. By leveraging violations of spatiotemporal scene consistency as a unified motion cue, the proposed approach achieves category-agnostic and temporally consistent tracking without resorting to heavyweight multi-stage pipelines or additional semantic priors. This lightweight yet effective formulation satisfies the latency and efficiency demands of real-time SLAM while preserving the continuity and completeness of dynamic object behavior across time.

\subsection{Camera Tracking}
\label{m3}
For each incoming frame, we first perform frame-to-model camera tracking. The pose $T_t$ is initialized using a constant-velocity motion model and then refined by aligning the rendered observations from the static Gaussian map with the incoming RGB-D frame.

To ensure that pose estimation is not contaminated by moving objects, we restrict optimization to reliable static regions. Specifically, we define the tracking mask
\begin{equation}
M_{track} = (\neg M_d) \cap (\hat{O} > \tau_{track}),
\end{equation}
where $\neg$ denotes mask negation. $M_d=\bigcup_{i \in \mathcal{D}} \{ M_i \, | \, a_i = 1 \}$ denotes the union of active dynamic masks identified by our CAD module, and $\hat{O}$ is the accumulated opacity of the rendered static map. The threshold $\tau_{track}$ filters out low-opacity areas that correspond to poorly reconstructed or sparsely observed regions.

Within this stable region, camera tracking is supervised by both photometric and geometric loss. 
\begin{equation}
L_I = \sum M_{track}\cdot\lVert \hat{I} - I \rVert_1,
\end{equation}

\begin{equation}
L_D = \sum (M_{track} \cap M_v)\cdot\lVert \hat{D} - D \rVert_1,
\end{equation}
where $M_v$ indicates the valid pixels of the input depth map, accounting for missing values or sensor-induced holes. 

The final camera pose is obtained by minimizing a weighted combination of the two terms:
\begin{equation}
T_t = \operatorname*{arg\,min}_{T} \ \lambda_{track} L_I + (1 - \lambda_{track}) L_D,
\end{equation}
where $\lambda_{track}$ balances the contribution of photometric and geometric cues. This formulation ensures that pose refinement is guided exclusively by reliable static observations, yielding robust tracking even in highly dynamic environments.

To mitigate pose drift in extended sequences, we incorporate loop detection and bundle adjustment (BA). Following Droid-SLAM \cite{teed2021droid}, we integrate a pre-trained optical flow model with Dense Bundle Adjustment Layer (DBA) to optimize keyframe camera poses and depth. In contrast to WildGS-SLAM \cite{zheng2025wildgs} which introduces uncertainty maps during BA optimization, we take advantage of acquired dynamic masks to eliminate interference from moving entities while preserving the integrity of the static scene. Following DG-SLAM \cite{xu2024dg}, the cost function over the keyframe graph is defined as:
\begin{equation}
\begin{aligned}
    \mathbf{E}(\mathbf{T}, \mathbf{d}) = &\sum_{(i,j)\in\mathcal{E}} \|\mathbf{p}_{ij}^* - \Pi_c(\mathbf{T}_{ij} \circ \Pi_c^{-1}(\mathbf{p}_i, \mathbf{d}_i))\|^2_{\Sigma_{ij} \cdot \neg M_d}, \\
    \Sigma_{ij} = &\operatorname{diag} \omega_{ij},
\end{aligned}
\end{equation}
where $\mathbf{p}_{ij}^*$
where $\Pi_c$ denotes the projection transformation from 3D coordinates to the image plane. $\mathbf{p}_i$ represents pixel coordinates, $\mathbf{d}_i$ indicates inverse depth values. $\mathbf{T}_{ij}$ corresponds to the relative camera pose between frames $i$ and $j$. $\mathbf{p}_{ij}^*$ represents the propagated coordinates of pixel $\mathbf{p}_i$ in frame $j$ through optical flow estimation. $\| \cdot \|^2_{\Sigma_{ij} \cdot \neg M_d}$ denotes the Mahalanobis distance weighted by confidence metric $\Sigma_{ij}$, while filtering the dynamic.


\subsection{Dynamic-static Decoupled Mapping}
\label{m4}

Unlike existing dynamic SLAM \cite{xu2024dg, zheng2025wildgs, wen2024gassidy} that perform elimination of dynamic objects, we propose a dynamic-static decoupled mapping strategy that leverages temporal Gaussian models to achieve online incremental dynamic modeling.

\noindent \textbf{Dynamic-Static Separation.}
At the beginning of SLAM, the scene is assumed to be entirely static, and the initial map is constructed accordingly. Once the CAD module detects a dynamic object, we immediately decouple it from the static map. Following Eq.~\ref{eq:mobile_sam}, we obtain a complete dynamic-object mask from the image using the inconsistency-guided prompt. With its 2D support region identified, the corresponding 3D points are recovered via back-projection using the depth map and camera intrinsics, yielding a point cloud of the dynamic object. We then extract all Gaussian primitives whose centers fall within this dynamic region from the static map, thereby achieving adaptive and precise dynamic–static separation.

\noindent \textbf{Static Mapping.}
For each newly added keyframe, we insert Gaussian ellipsoids into the static Gaussian map to fill regions with holes or poor quality in the rendered output.
\begin{equation}
M_{instert} = (\neg M_d) \cap M_v \cap (\hat{O} < \tau_{map}),
\end{equation}
where $\tau_{map}$ represents the opacity threshold for the static mapping region. The RGB-D pixels in $M_{insert}$ will be initialized as Gaussian ellipsoids and incorporated into the static Gaussian map. Subsequently, keyframes are selected from the keyframe set to optimize the static map. We compute the rendering loss for the valid static regions:
\begin{equation}
\begin{split}
L_I = (1-\lambda_{ssim}) \frac{1}{M_{map}} \sum M_{map} \cdot \lVert \hat{I} - I\rVert_1 
\\
+ \lambda_{ssim} SSIM(\hat{I}, I,  M_{map}),
\end{split}
\end{equation}
\begin{equation}
L_D = \frac{1}{M_{map}} \sum M_{map} \cdot \lVert \hat{D} - D\rVert_1, 
\end{equation}
\begin{equation}
M_{map} = (\neg M_d) \cap M_v,
\end{equation}

where $\lambda_{ssim}$ is the weight of ssim loss. 
The final static mapping loss is:
\begin{equation}
L_{map}^{static} = \lambda_{color} L_{I} + \lambda_{depth}L_{D} + \lambda_{reg}L_{reg},
\end{equation}
where $L_{reg}$ is the Gaussian ellipsoid scale regularization loss from \cite{zhu2024loopsplat}.


\noindent \textbf{Dynamic Mapping.}
For each tracked dynamic object $i \in \mathcal{D}$, we construct a temporal Gaussian model:
\begin{equation}
\boldsymbol{G}^{i}(t) = \{(\boldsymbol{\mu}_j^t,\boldsymbol{\Sigma}_j^t, o_j^t, \boldsymbol{h}_j^t)\}.
\end{equation}

$\boldsymbol{G}^{i}(t)$ is initialized with point cloud back-projected from $I_t,D_t$ using the mask $M_i^t$. Then the dynamic object $i$ at $t$ is rendered to yield the predicted color $\hat{I}_i^t$ and depth $\hat{D}_i^t$:
\begin{equation}
   \hat{I}^t_i = \sum_{j \in \boldsymbol{G}^{i}(t)}c_j^t \alpha_j^t \prod_{l=1}^{j-1}(1-\alpha_l^t),
\end{equation}
\begin{equation}
   \hat{D}^t_i = \sum_{j \in \boldsymbol{G}^{i}(t)}d_j^t \alpha_j^t \prod_{l=1}^{j-1}(1-\alpha_l^t).
\end{equation}

The parameters of the dynamic Gaussians $\boldsymbol{G}^{i}(t)$ are optimized using loss defined over the valid mapping region $M_{\text{map}}^{i}=M^i \cap M_v$. Here we omit $t$ for clarity. 

\begin{equation}
\begin{split}
L_I^{i} = (1-\lambda_{ssim}) \frac{1}{M_{map}^{i}} \sum M_{map}^{i} \cdot \lVert \hat{I}_d - I\rVert_1 
\\+ \lambda_{ssim} SSIM(\hat{I}_i, I,  M_{map}^{i}),
\end{split}
\end{equation}

\begin{equation}
L_D^{i} = \frac{1}{M_{map}^{i}} \sum M_{map}^{i} \cdot \lVert \hat{D}_i - D\rVert_1.
\end{equation}

The final dynamic mapping loss of object $i$ is:
\begin{equation}
L_{map}^{i} = \lambda_{color} L_{I}^i + \lambda_{depth}L_{D}^i + \lambda_{reg}L_{reg}.
\end{equation}

\begin{figure*}[t] 
\center{\includegraphics[width=1.0\textwidth]{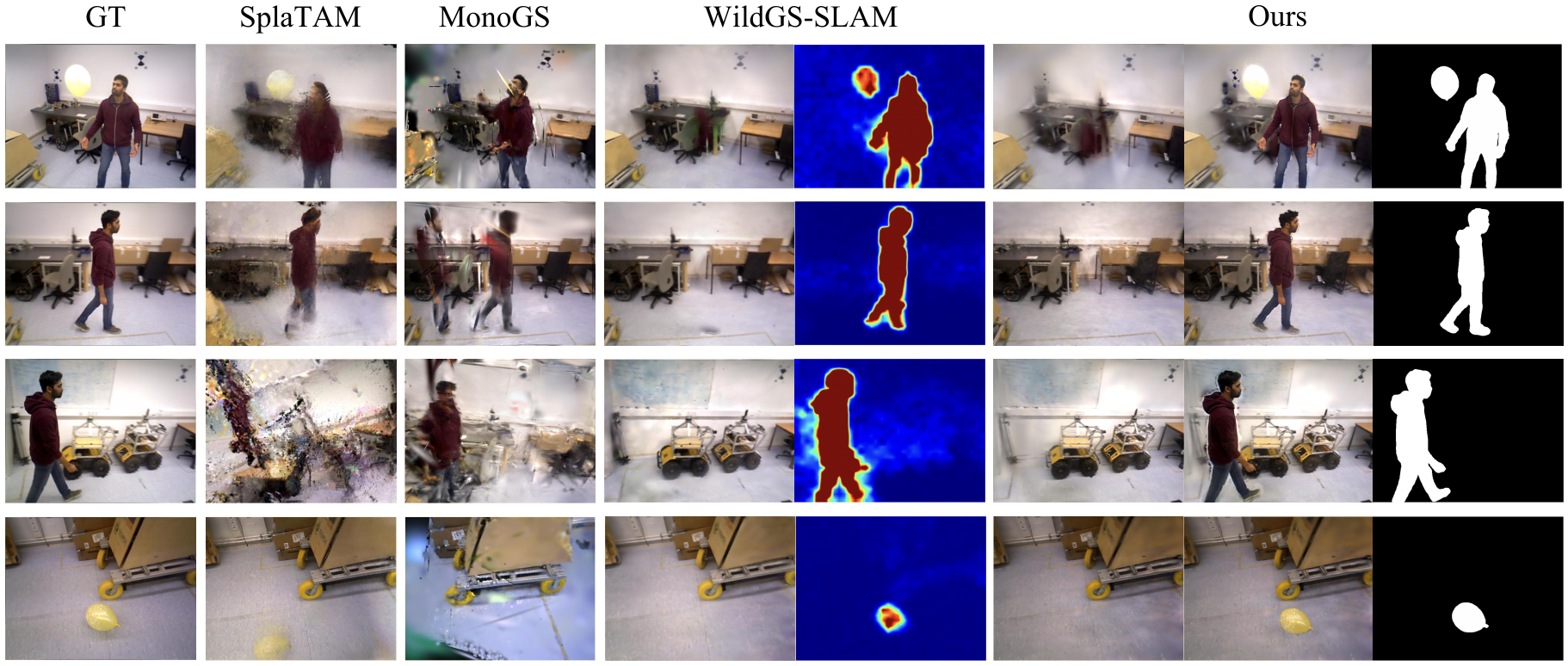}} 
\caption{Rendering visualization. SLAM methods assuming a static world fail to reconstruct dynamic scenes, highlighting the necessity of dynamic-object identification. Although uncertainty-based method can detect motion to some extent, they still leave residual artifacts of dynamic objects (red circles in the left column) and introduce blurring in static regions (right column). In contrast, CAD-SLAM realizes dynamic-static decoupled mapping, demonstrating superior mapping performance in highly dynamic environments.
}
\label{fig_bonn_render}

\end{figure*}

\begin{figure*}[!t] 
\center{\includegraphics[width=1.0\textwidth]{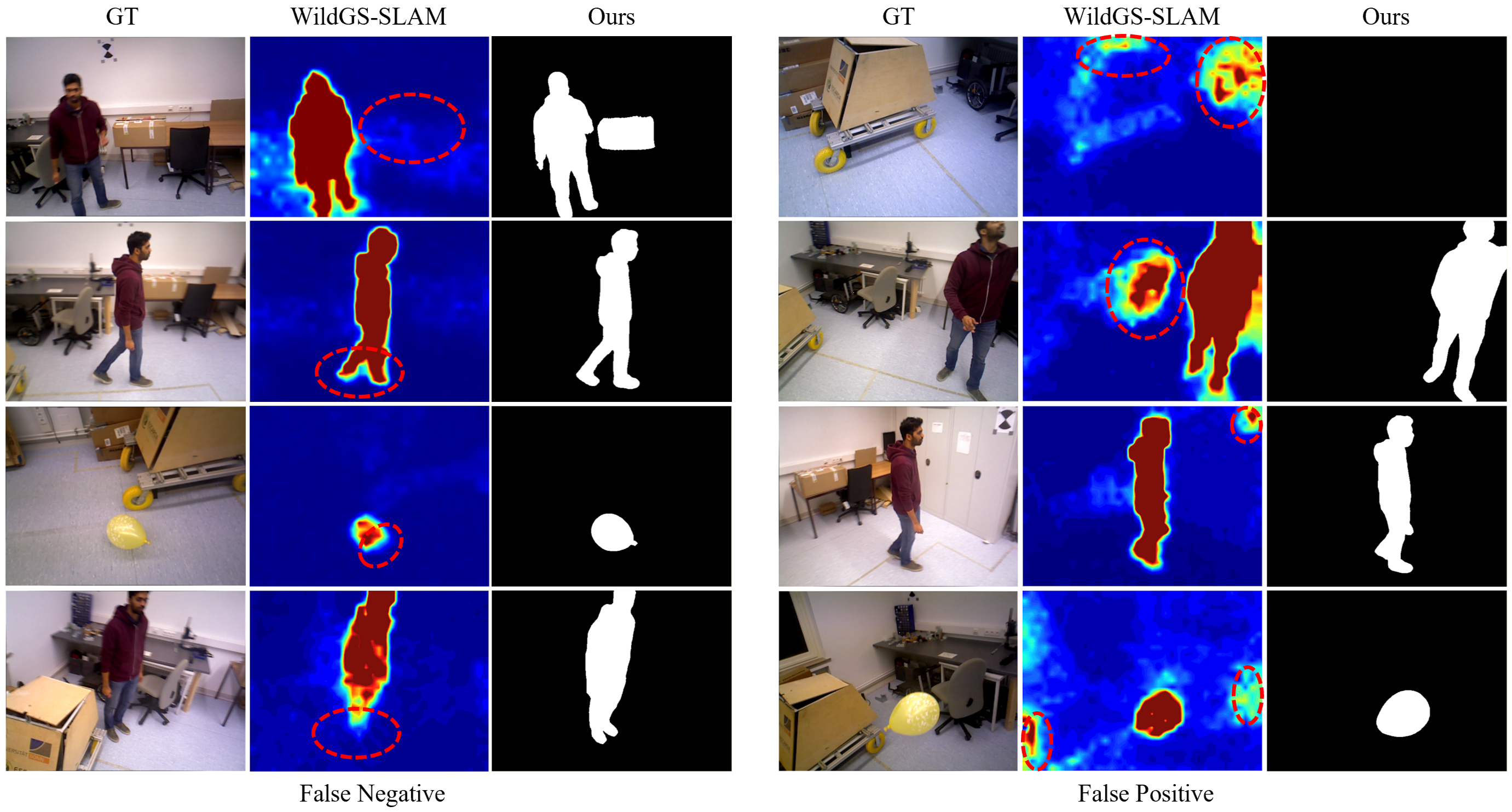}} 
\caption{
Comparison between the uncertainty maps of WildGS-SLAM~\cite{zheng2025wildgs} and the dynamic masks produced by CAD-SLAM (Ours). Yellow regions denote true dynamic objects, while red circles highlight uncertainty failure cases. As shown, uncertainty-based detection exhibits notable false negatives—either missing entire dynamic objects or only partially segmenting them. It also suffers from false positives, where high uncertainty incorrectly appears in static regions with sparse observations or temporary occlusions. In contrast, our dynamic masks produced by CAD-model are relatively robust and accurate.
}
\label{fig_bonn_mask_wild}
\end{figure*}

\begin{table*}[!t]
\caption{Camera tracking results on Bonn dataset. "*" denotes the version reproduced by NICE-SLAM. "-" denotes the absence of mention. "X" denotes a failure in execution, with no valid result. "$\dag$" indicates the replacement of depth estimation with ground-truth depth for fair comparison. The metric unit is [cm]. 
 Best results are highlighted as \colorbox{tabfirst}{first}, \colorbox{tabsecond}{second}, and \colorbox{tabthird}{third}.}
\label{tab_bonn_tracking}
\centering
\resizebox{\textwidth}{!}{
\begin{tabular}{cccccccccccccccc}
\toprule
\multicolumn{1}{c}{Methods}  & \multicolumn{2}{c}{\texttt{balloon}} & \multicolumn{2}{c}{\texttt{balloon2}} & \multicolumn{2}{c}{\texttt{ps\_track}}& \multicolumn{2}{c}{\texttt{ps\_track2}}&\multicolumn{2}{c}{\texttt{ball\_track}}& \multicolumn{2}{c}{\texttt{mv\_box2}}& \multicolumn{2}{c}{Avg.}\\
           
\midrule

\multicolumn{1}{l}{\textit{Traditional}}&\textit{RMSE}&\textit{S.D.}&\textit{RMSE}&\textit{S.D.}&\textit{RMSE}&\textit{S.D.}&\textit{RMSE}&\textit{S.D.}&\textit{RMSE}&\textit{S.D.}&\textit{RMSE}&\textit{S.D.}&\textit{RMSE}&\textit{S.D.}\\

\multicolumn{1}{l}{ORB-SLAM3\cite{campos2021orb}} &5.8 &2.8 &17.7 &8.6 &70.7 &32.6 &77.9 &43.8 & \cellcolor{tabfirst}3.1 & \cellcolor{tabsecond}1.6 &{3.5} &\cellcolor{tabthird}1.5 &29.8 &15.2 \\
\multicolumn{1}{l}{Droid-VO\cite{teed2021droid}} & {5.4}& -& {4.6}&- & 21.34&- &46.0&- &8.9&-&{5.9}&- &15.4&- \\
\multicolumn{1}{l}{DynaSLAM\cite{bescos2018dynaslam}} &3.0&-&2.9&-&6.1&-&7.8&-&4.9&-&3.9&-&4.77&-\\
\multicolumn{1}{l}{ReFusion\cite{palazzolo2019refusion}} &17.5&-&25.4&-&28.9&-&46.3&-&30.2&-&17.9&-&27.7&-\\
\midrule

\multicolumn{1}{l}{\textit{NeRF based}}&\textit{RMSE}&\textit{S.D.}&\textit{RMSE}&\textit{S.D.}&\textit{RMSE}&\textit{S.D.}&\textit{RMSE}&\textit{S.D.}&\textit{RMSE}&\textit{S.D.}&\textit{RMSE}&\textit{S.D.}&\textit{RMSE}&\textit{S.D.}\\

\multicolumn{1}{l}{iMAP*\cite{sucar2021imap}} &14.9&5.4&67.0&19.2&28.3&12.9&52.8&20.9&24.8&11.2&28.3&35.3&36.1&17.5\\
\multicolumn{1}{l}{NICE-SLAM\cite{zhu2022nice}} &X&X&66.8&20.0&54.9&27.5&45.3&17.5&21.2&13.1&31.9&13.6& - & - \\
\multicolumn{1}{l}{Vox-Fusion\cite{yang2022vox}}  &65.7 &30.9 &82.1 &52.0 &128.6 &52.5 &162.2 &46.2 &43.9 &16.5 &47.5 &19.5 &88.4 &36.3\\
\multicolumn{1}{l}{Co-SLAM\cite{wang2023co}} &28.8 &9.6 &20.6 &8.1 &61.0 &22.2 &59.1 &24.0 &38.3 &17.4 &70.0 &25.5 &46.3 &17.8\\
\multicolumn{1}{l}{ESLAM\cite{{johari2023eslam}}} &22.6 &12.2 &36.2 &19.9 &48.0 &18.7 &51.4 &23.2 &12.4 &6.6 &17.7 &7.5 &31.4 &14.7\\
\multicolumn{1}{l}{RoDyn-SLAM\cite{jiang2024rodyn}} &7.9 &2.7&11.5 &6.1 &{14.5} &\cellcolor{tabthird}4.6 &{13.8} &\cellcolor{tabthird}3.5 &13.3 &4.7 &12.6 &4.7 &12.3 &4.38\\
\multicolumn{1}{l}{DynaMoN(MS\&SS)\cite{schischka2024dynamon}} &\cellcolor{tabthird}2.8 &- &2.7 &- &14.8 &- &\cellcolor{tabfirst}2.2 &- &\cellcolor{tabsecond}{3.4}&- &2.7&- &{4.77} &-\\
\midrule

\multicolumn{1}{l}{\textit{3DGS based}}&\textit{RMSE}&\textit{S.D.}&\textit{RMSE}&\textit{S.D.}&\textit{RMSE}&\textit{S.D.}&\textit{RMSE}&\textit{S.D.}&\textit{RMSE}&\textit{S.D.}&\textit{RMSE}&\textit{S.D.}&\textit{RMSE}&\textit{S.D.}\\
\multicolumn{1}{l}{SplaTAM\cite{keetha2024splatam}} &40.0 &14.6 	&39.5 	&15.8	&217.9 &81.2 	&131.0 &33.1 	&20.2 	&16.3 	&17.1 	&9.3 	&77.6 	&28.4 \\
\multicolumn{1}{l}{MonoGS\cite{Matsuki:Murai:etal:CVPR2024}} &31.2 &15.3 	&26.7 	&13.5	&43.8 &16.8 	&48.4 &16.6 	&\cellcolor{tabthird}{4.7}	&{2.4}	&7.1 &3.5	&27.0 	&11.4 \\
\multicolumn{1}{l}{GS-ICP SLAM\cite{ha2025rgbd}} & 42.2 & 14.4 & 57.5 & 22.4 & 87.8 & 40.6 & 49.8 & 21.2 & 32.1 & 11.7 & 26.0 & 12.4 & 49.2 & 20.5 \\

\multicolumn{1}{l}{PG-SLAM\cite{li2025pg}} & 6.4 &2.2 &7.3 &\cellcolor{tabthird} 3.4 &5.0 &\cellcolor{tabsecond} 1.9 & 8.5 & \cellcolor{tabsecond}2.8 & - &-& 7.0& 2.0 & -& -\\

\multicolumn{1}{l}{DG-SLAM\cite{xu2024dg}} &  3.7 & - & 4.1 & - & 4.5 & - & {6.9} & - & 10.0 & - & {3.5} & - & 5.45 &- \\

\multicolumn{1}{l} {WildGS-SLAM\cite{zheng2025wildgs}}  & 2.9 &\cellcolor{tabthird}1.2 & \cellcolor{tabthird}2.5 &\cellcolor{tabsecond}1.2 & \cellcolor{tabthird}3.6 & \cellcolor{tabsecond}1.9 & \cellcolor{tabsecond}3.1 & \cellcolor{tabfirst}1.4 & \cellcolor{tabfirst}3.1 & \cellcolor{tabsecond}1.6 & \cellcolor{tabsecond}2.4 & \cellcolor{tabsecond}1.3 & \cellcolor{tabthird}2.93 & \cellcolor{tabsecond}1.43 \\
\multicolumn{1}{l} {WildGS-SLAM\dag\cite{zheng2025wildgs}}  &  \cellcolor{tabfirst}2.4 & \cellcolor{tabfirst}1.0 & \cellcolor{tabsecond}2.4 & \cellcolor{tabsecond}1.2 & \cellcolor{tabsecond}{3.4} & \cellcolor{tabsecond}1.9 & \cellcolor{tabsecond}3.1 & \cellcolor{tabfirst}1.4 & \cellcolor{tabsecond}3.4 & \cellcolor{tabthird}1.9 & \cellcolor{tabthird}2.6 & \cellcolor{tabsecond}1.3 & \cellcolor{tabsecond}2.88 &\cellcolor{tabthird}1.45 \\

\multicolumn{1}{l}{CAD-SLAM(Ours)} & \cellcolor{tabsecond}2.7&\cellcolor{tabsecond} 1.1 & \cellcolor{tabfirst}2.3 & \cellcolor{tabfirst}0.8 & \cellcolor{tabfirst}2.4 & \cellcolor{tabfirst}1.1 & \cellcolor{tabthird}3.7 & \cellcolor{tabfirst}1.4& \cellcolor{tabsecond}3.4 & \cellcolor{tabfirst}1.2& \cellcolor{tabfirst}2.1 & \cellcolor{tabfirst}0.8 & 
  \cellcolor{tabfirst}2.77 &  \cellcolor{tabfirst} 1.05 \\
\bottomrule
\end{tabular}}

\end{table*}

\begin{table*}[h]
\caption{Camera tracking results on TUM RGB-D dataset."*" denotes the version reproduced by NICE-SLAM. "-" denotes the absence of mention. "X" denotes a failure in execution, with no valid result. "$\dag$" indicates the replacement of depth estimation with ground-truth depth for fair comparison. The metric unit is [cm]. 
 Best results are highlighted as \colorbox{tabfirst}{first}, \colorbox{tabsecond}{second}, and \colorbox{tabthird}{third}.}
\label{tab_tum_tracking}
\centering
\resizebox{\textwidth}{!}{
\begin{tabular}{cccccccccccccccc}
\toprule
\multirow{2}{*}{Methods} &\multicolumn{8}{c}{Dynamic}  &\multicolumn{4}{c}{Static}&\multicolumn{2}{c}{\multirow{2}{*}{Avg.}} \\
&\multicolumn{2}{c}{\texttt{fr3/wk\_xyz}}   & \multicolumn{2}{c}{\texttt{fr3/wk\_hf}}  & \multicolumn{2}{c}{\texttt{fr3/wk\_st}} &\multicolumn{2}{c}{\texttt{fr3/st\_hf}}&\multicolumn{2}{c}{\texttt{fr1/xyz}} & \multicolumn{2}{c}{\texttt{fr1/rpy}} & \\                   
\midrule

\multicolumn{1}{l}{\textit{Traditional}}&\textit{RMSE}&\textit{S.D.}&\textit{RMSE}&\textit{S.D.}&\textit{RMSE}&\textit{S.D.}&\textit{RMSE}&\textit{S.D.}&\textit{RMSE}&\textit{S.D.}&\textit{RMSE}&\textit{S.D.}&\textit{RMSE}&\textit{S.D.}\\

\multicolumn{1}{l}{ORB-SLAM3 \cite{campos2021orb}} &28.1&12.2&30.5&9.0&2.0&1.1&2.6&1.6&\cellcolor{tabthird}1.1&\cellcolor{tabthird}0.6&\cellcolor{tabthird}2.2&1.3&11.1&4.3 \\
\multicolumn{1}{l}{DVO-SLAM\cite{kerl2013dense}} &59.7&-&52.9&-&21.2&-&6.2&-&\cellcolor{tabthird}1.1&-&\cellcolor{tabsecond}2.0&-&22.9&-\\
\multicolumn{1}{l}{DynaSLAM\cite{bescos2018dynaslam}} &1.7&-&2.6&-&\cellcolor{tabsecond}0.7&-&2.8&-&-&-&-&-&-&-\\
\multicolumn{1}{l}{ReFusion\cite{palazzolo2019refusion}} &9.9&-&10.4&-&1.7&-&11.0&-&-&-&-&-&-&-\\

\midrule

\multicolumn{1}{l}{\textit{NeRF based}}&\textit{RMSE}&\textit{S.D.}&\textit{RMSE}&\textit{S.D.}&\textit{RMSE}&\textit{S.D.}&\textit{RMSE}&\textit{S.D.}&\textit{RMSE}&\textit{S.D.}&\textit{RMSE}&\textit{S.D.}&\textit{RMSE}&\textit{S.D.}\\

\multicolumn{1}{l}{iMAP*\cite{sucar2021imap}} &111.5&43.9&X&X&137.3&21.7&93.0&35.3&7.9&7.3&16.0&13.8& - & - \\
\multicolumn{1}{l}{NICE-SLAM\cite{zhu2022nice}} &113.8&42.9&X&X&88.2&27.8&45.0&14.4&4.6&3.8&3.4&2.5& - & - \\
\multicolumn{1}{l}{Vox-Fusion\cite{yang2022vox}} &146.6&32.1&X&X&109.9&25.5&89.1&28.5&1.8&0.9&4.3&3.0& - & - \\
\multicolumn{1}{l}{Co-SLAM\cite{wang2023co}} &51.8&25.3&105.1&42.0&49.5&10.8&4.7&{2.2}&2.3&1.2&3.9&2.8&36.3&14.1\\
\multicolumn{1}{l}{ESLAM\cite{johari2023eslam}} &45.7&28.5&60.8&27.9&93.6&20.7&3.6&{1.6}&\cellcolor{tabthird}1.1&\cellcolor{tabthird}0.6&\cellcolor{tabthird}2.2&\cellcolor{tabthird}1.2&34.5&13.5\\
\multicolumn{1}{l}{RoDyn-SLAM\cite{jiang2024rodyn}} &8.3 & 5.5 &5.6 & \cellcolor{tabthird}2.8 & 1.7  & \cellcolor{tabthird}0.9& 4.4 & 2.2 & 1.5 & 0.8 & 2.8 & 1.5&4.05 & 2.28\\
\multicolumn{1}{l}{DynaMoN(MS\&SS)\cite{schischka2024dynamon}}  &\cellcolor{tabsecond}1.4 & - & 1.9 & - & \cellcolor{tabsecond}0.7 & - & 2.3& - & - & - & - & - &- &- & - \\
\midrule

\multicolumn{1}{l}{\textit{3DGS based}}&\textit{RMSE}&\textit{S.D.}&\textit{RMSE}&\textit{S.D.}&\textit{RMSE}&\textit{S.D.}&\textit{RMSE}&\textit{S.D.}&\textit{RMSE}&\textit{S.D.}&\textit{RMSE}&\textit{S.D.}&\textit{RMSE}&\textit{S.D.}\\
\multicolumn{1}{l}{SplaTAM\cite{keetha2024splatam}}&160.5&42.4&X&X&42.6&13.0&14.5&6.2&\cellcolor{tabthird}{1.1}&\cellcolor{tabthird}{0.6}&3.2&1.5&44.4&12.8\\
\multicolumn{1}{l}{MonoGS\cite{Matsuki:Murai:etal:CVPR2024}} &28.4&12.3&47.8&16.5&15.3&8.4&13.9&3.2&\cellcolor{tabsecond}1.0&\cellcolor{tabfirst}0.4&2.6&1.3	&18.2&7.0\\
\multicolumn{1}{l}{GS-ICP\cite{ha2025rgbd}} &68.9 & 50.8 & 84.6 & 34.3 & 87.5 & 16.9 & 11.2 & 2.7 & 1.4 & 0.7 & 4.2 & 3.9 & 43.0 & 18.2 \\
\multicolumn{1}{l}{PG-SLAM\cite{li2025pg}} &6.8 &\cellcolor{tabthird} 2.9 & 11.7 &4.4&\cellcolor{tabthird}1.4 &\cellcolor{tabsecond}0.6 &  4.0 & 1.5 &- & - & - &- &-\\

\multicolumn{1}{l}{DG-SLAM\cite{xu2024dg}} &\cellcolor{tabthird}1.6 &- & \cellcolor{tabfirst}0.6& - & - & - & -& - & - & - & - & - &- &-   \\
\multicolumn{1}{l} {WildGS-SLAM\cite{zheng2025wildgs}}  & \cellcolor{tabfirst}1.2&\cellcolor{tabfirst}0.6 & \cellcolor{tabthird}1.5& \cellcolor{tabsecond}0.8 &\cellcolor{tabfirst}0.5 &\cellcolor{tabfirst}0.2& \cellcolor{tabthird}1.8 &\cellcolor{tabthird}1.1&\cellcolor{tabfirst}0.9&\cellcolor{tabsecond}0.5&2.3&1.3&\cellcolor{tabthird}1.37&\cellcolor{tabthird}0.75  \\
\multicolumn{1}{l} {WildGS-SLAM\dag\cite{zheng2025wildgs}}  & \cellcolor{tabfirst} 1.2 &\cellcolor{tabfirst}0.6&\cellcolor{tabsecond}1.4 &\cellcolor{tabfirst}0.7 & \cellcolor{tabfirst}0.5& \cellcolor{tabfirst}0.2& \cellcolor{tabsecond}1.7 &\cellcolor{tabsecond}1.0 &\cellcolor{tabfirst}0.9&\cellcolor{tabsecond}0.5 &\cellcolor{tabthird}2.2&\cellcolor{tabsecond}1.0 &\cellcolor{tabsecond}1.32& \cellcolor{tabsecond}0.67 \\

\multicolumn{1}{l}{CAD-SLAM(Ours)} & \cellcolor{tabsecond}1.4 & \cellcolor{tabsecond}0.9 &1.6 & \cellcolor{tabsecond}0.8 & \cellcolor{tabfirst}0.5 & \cellcolor{tabfirst}0.2 & \cellcolor{tabfirst}1.3& \cellcolor{tabfirst}0.6 & \cellcolor{tabsecond}1.0 & \cellcolor{tabsecond}0.5 & \cellcolor{tabfirst}1.7 & \cellcolor{tabfirst}0.9 &\cellcolor{tabfirst}1.25 & \cellcolor{tabfirst}0.65 \\

\bottomrule
\end{tabular}}

\end{table*}

\begin{figure}[!t] 
\center{\includegraphics[width=1.0\linewidth]{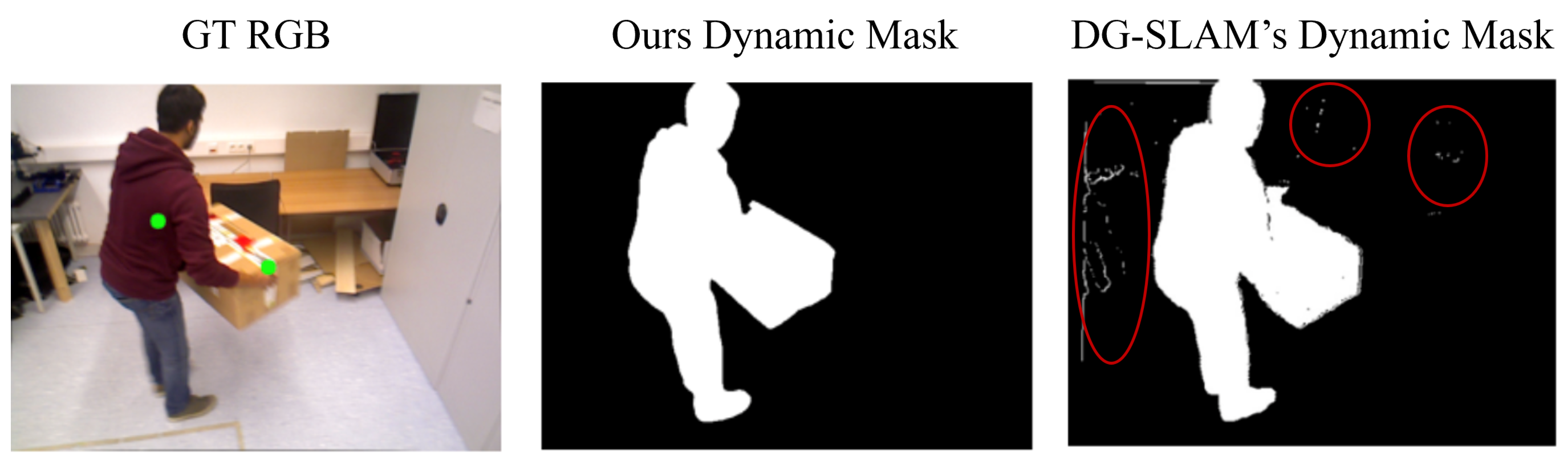}} 
\caption{Comparison of the masks produced by our method and DG‑SLAM. Red circles highlight the noise artifacts in DG‑SLAM’s mask, which caused by inaccurate depth warping.}
\label{fig_bonn_mask_dg}
\end{figure}

\begin{table*}[t]
\centering
\caption{Camera tracking results on Wild-SLAM MoCap Dataset (ATE RMSE $\downarrow$ [cm]). "X" denotes a failure in execution, with no valid result. Best results are highlighted as \colorbox{tabfirst}{first}, \colorbox{tabsecond}{second}, and \colorbox{tabthird}{third}.}
\label{tab_wild_tracking}

\resizebox{\linewidth}{!}{ 
\begin{tabular}{l c c c c c c c c c c c}
\toprule
Method & ANymal1 & ANymal2 & Ball & Crowd & Person & Racket & Stones & Table1 & Table2 & Umbrella & Avg \\
\midrule
ReFusion \cite{palazzolo2019refusion} 
& 4.2 & 5.6 & 5.0 & 91.9 & 5.0 & 10.4 & 39.4 & 99.1 & 101.0 & 10.7 & 37.23 \\

DynaSLAM (N+G) \cite{bescos2018dynaslam} 
& 1.6 & \cellcolor{tabthird}0.5 & \cellcolor{tabthird}0.5 
& 1.7 
& \cellcolor{tabsecond}0.5 
& 0.8 
& 2.1 
& \cellcolor{tabthird}1.2 
& 34.8 
& 34.7 
& 7.84 \\

NICE-SLAM \cite{zhu2022nice} 
& X & 123.6 & 21.1 & X & 150.2 & X & 134.4 & 138.4 & X & 23.8 & - \\

DSO \cite{DSO} 
& 12.0 & 2.5 & 1.0 & 88.6 & 9.3 & 3.1 & 41.5 & 50.6 & 85.3 & 26.0 & 32.99 \\

DROID-SLAM \cite{teed2021droid} 
& \cellcolor{tabthird}0.6 & 4.7 & 1.2 
& 2.3 
& \cellcolor{tabthird}0.6 
& 1.5 
& 3.4 
& 48.0 
& 95.6 
& 3.8 
& 16.17 \\

DynaSLAM (RGB) \cite{bescos2018dynaslam} 
& \cellcolor{tabthird}0.6 & \cellcolor{tabthird}0.5 & \cellcolor{tabthird}0.5 
& \cellcolor{tabsecond}0.5 
& \cellcolor{tabfirst}0.4 
& \cellcolor{tabthird}0.6 
& \cellcolor{tabthird}1.7 
& 1.8 
& 42.1 
& 1.2 
& 5.19 \\

MonoGS \cite{Matsuki:Murai:etal:CVPR2024} 
& 8.8 & 51.6 & 7.4 & 70.3 & 55.6 & 67.6 & 39.9 & 24.9 & 118.4 & 35.3 & 47.99 \\

SplatGS \cite{sandstrom2025splat} 
& \cellcolor{tabsecond}0.4 & \cellcolor{tabsecond}0.4 & \cellcolor{tabsecond}0.3 
& \cellcolor{tabthird}0.7 
& 0.8 
& \cellcolor{tabthird}0.6 
& 1.9 
& 2.5 
& 73.6 
& 5.9 
& 8.71 \\

MonST3R-SW \cite{zhang2024monst3r} 
& 3.5 & 21.6 & 6.1 & 14.4 & 7.2 & 13.2 & 11.2 & 4.8 & 33.7 & 5.5 & 12.12 \\

MegaSAM \cite{li2025megasam} 
& \cellcolor{tabthird}0.6 & 2.7 & 0.6 
& 1.0 
& 3.2 
& 1.6 
& 3.2 
& \cellcolor{tabsecond}1.0 
& \cellcolor{tabthird}9.4 
& \cellcolor{tabthird}0.6 
& \cellcolor{tabthird}2.40 \\

WildGS-SLAM~\cite{zheng2025wildgs} 
& \cellcolor{tabfirst}0.2 & \cellcolor{tabfirst}0.3 & \cellcolor{tabfirst}0.2 
& \cellcolor{tabfirst}0.3 
& 0.8 
& \cellcolor{tabfirst}0.4 
& \cellcolor{tabfirst}0.3 
& \cellcolor{tabfirst}0.6 
& \cellcolor{tabfirst}1.3 
& \cellcolor{tabfirst}0.2 
& \cellcolor{tabfirst}0.46 \\

CAD-SLAM(Ours)
& \cellcolor{tabsecond}0.4 & \cellcolor{tabfirst}0.3 & \cellcolor{tabfirst}0.2 
& 1.5 
& 0.8 
& \cellcolor{tabsecond}0.5 
& \cellcolor{tabsecond}0.7 
& \cellcolor{tabfirst}0.6 
& \cellcolor{tabsecond}1.5 
& \cellcolor{tabsecond}0.4 
& \cellcolor{tabsecond}0.68 \\
\bottomrule
\end{tabular}
}
\end{table*}

\begin{figure*}[!t] 
\center{\includegraphics[width=1.0\textwidth]{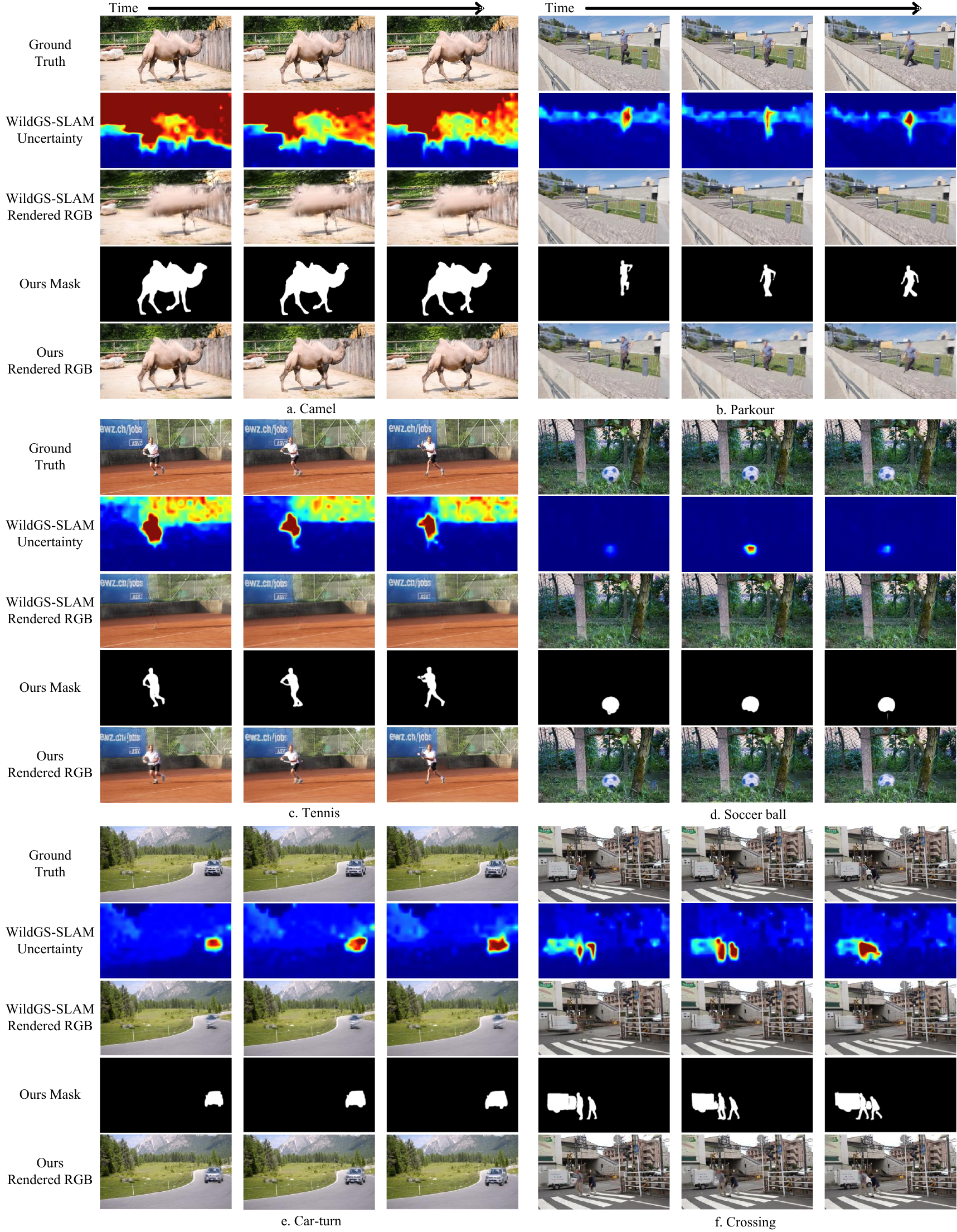}} 
\vspace{-22pt}
\caption{
Visualization of dynamic mask and rendering results on the DAVIS Dataset. Our dynamic masks are more complete and precise than the uncertainty masks of WildGS-SLAM~\cite{zheng2025wildgs} and remain unaffected by background interference. Due to the failure of uncertainty, WildGS-SLAM results in blurriness in the dynamic area. In contrast, our method can accurately identify and model various types of dynamic objects, shown ours stronger adaptability.
}
\label{fig_davis_render}
\end{figure*}

\begin{table*}[htbp]
  \centering
  \caption{Reconstruction results on dynamic scene sequences in the BONN dataset. Instances of tracking failures are denoted by "X". Best results are highlighted as \colorbox{tabfirst}{first} and \colorbox{tabsecond}{second}}
  \label{tab_bonn_3d}
\resizebox{0.85\textwidth}{!}{
  \begin{tabular}{llcccccc}
    \toprule
    Method             & Metric                          & ball      & ball2     & ps\_trk    & ps\_trk2   & mv\_box2   & Avg.      \\
    \midrule
    \multirow{3}{*}{NICE-SLAM~\cite{zhu2022nice}}
                      & Acc. [cm] $\downarrow$          & X         & 24.30     & 43.11     & 74.92     & 17.56     & 39.97     \\
                      & Comp. [cm] $\downarrow$         & X         & 16.65     & 117.95    & 172.20    & 18.19     & 81.25     \\
                      & Comp. Ratio [$\leq$5cm\%] $\uparrow$ & X     & 29.68     & 15.89     & 13.96     & 32.18     & 22.93     \\
    \midrule
    \multirow{3}{*}{Co-SLAM~\cite{wang2023co}}
                      & Acc. [cm] $\downarrow$          & 10.61     & 14.49     & 26.46     & 26.00     & 12.73     & 18.06     \\
                      & Comp. [cm] $\downarrow$         & 10.65     & 40.23     & 124.86    & 118.35    & 10.22     & 60.86     \\
                      & Comp. Ratio [$\leq$5cm\%] $\uparrow$ & 34.10  & 3.21      & 2.05      & 2.90      & 39.10     & 16.27     \\
    \midrule
    \multirow{3}{*}{ESLAM~\cite{johari2023eslam}}
                      & Acc. [cm] $\downarrow$          & 17.17     & 26.82     & 59.18     & 89.22     & 12.32     & 40.94     \\
                      & Comp. [cm] $\downarrow$         & \cellcolor{tabsecond}9.11      & 13.58     & 145.78    & 186.65    & 10.03     & 73.03     \\
                      & Comp. Ratio [$\leq$5cm\%] $\uparrow$ & 47.44  & 47.94     & 20.53     & 17.33     & 41.41     & 34.93     \\
    \midrule
    \multirow{3}{*}{DG-SLAM~\cite{xu2024dg}}
                      & Acc. [cm] $\downarrow$          & 7.00      & \cellcolor{tabfirst}{5.80}  & \cellcolor{tabsecond}9.14     & \cellcolor{tabsecond}11.78     & \cellcolor{tabsecond}6.56      & \cellcolor{tabsecond}8.06      \\
                      & Comp. [cm] $\downarrow$         &\cellcolor{tabsecond} 9.80      & \cellcolor{tabsecond}8.05      & \cellcolor{tabsecond}17.99     & \cellcolor{tabsecond}20.10     & \cellcolor{tabsecond}7.61      & \cellcolor{tabsecond}15.46     \\
                      & Comp. Ratio [$\leq$5cm\%] $\uparrow$ & \cellcolor{tabfirst}{49.46} & \cellcolor{tabsecond}52.41  & \cellcolor{tabsecond}34.62     & \cellcolor{tabsecond}32.81     & \cellcolor{tabfirst}{49.02} & \cellcolor{tabsecond}43.67  \\
    \midrule
    \multirow{3}{*}{CAD-SLAM(Ours)}
                      & Acc. [cm] $\downarrow$          & \cellcolor{tabfirst}{6.01}  & \cellcolor{tabsecond}7.21      & \cellcolor{tabfirst}{6.12}  & \cellcolor{tabfirst}{6.28}  & \cellcolor{tabfirst}{5.86}  & \cellcolor{tabfirst}{6.30}  \\
                      & Comp. [cm] $\downarrow$         & \cellcolor{tabfirst}{7.84}  & \cellcolor{tabfirst}{6.40}  & \cellcolor{tabfirst}{8.42}  & \cellcolor{tabfirst}{8.78}  & \cellcolor{tabfirst}{6.77}  & \cellcolor{tabfirst}{7.64}  \\
                      & Comp. Ratio [$\leq$5cm\%] $\uparrow$ & \cellcolor{tabsecond}45.82  & \cellcolor{tabfirst}{57.34} & \cellcolor{tabfirst}{40.72} & \cellcolor{tabfirst}{42.58} & \cellcolor{tabsecond}47.75  & \cellcolor{tabfirst}{46.84} \\
    \bottomrule
  \end{tabular}}
\end{table*}

\section{Experiment}
\label{sec:experiment}

\subsection{Experimental Setting}
\label{sec:exp_set}
\subsubsection{Dataset}
We evaluate CAD-SLAM on four real-world dynamic datasets: the TUM RGB-D~\cite{sturm2012benchmark} dataset, the Bonn dataset~\cite{palazzolo2019refusion}, the Wild-SLAM Dataset~\cite{zheng2025wildgs}, and the DAVIS dataset~\cite{perazzi2016benchmark}. 

\begin{itemize}
    \item \textbf{TUM RGB-D dataset}~\cite{sturm2012benchmark} offers a comprehensive collection of indoor RGB-D sequences recorded using a Microsoft Kinect sensor at 30 Hz with a resolution of 640×480. Ground-truth trajectories were captured using a high-precision motion capture system operating at 100 Hz. We select 4 dynamic sequences and 2 static  sequences for evaluation, covering diverse human motion patterns.
\item \textbf{Bonn dataset}~\cite{palazzolo2019refusion} is a dataset for RGB-D SLAM, containing highly dynamic sequences capturing human activities such as object manipulation and interaction with balloons. Each sequence includes ground-truth camera poses obtained via an OptiTrack Prime 13 motion capture system. We select 6 representative dynamic sequences for evaluation, including interactions between human and various objects.
\item \textbf{Wild-SLAM Dataset}~\cite{zheng2025wildgs} is primarily designed for benchmarking dynamic SLAM performance in unconstrained real-world environments. It incorporates multiple moving objects as dynamic interference sources to simulate common challenges such as motion-induced disturbances and occlusions encountered in practical scenarios. The data were captured using an Intel RealSense D455 camera. We evaluate on 10 core sequences.
\item \textbf{DAVIS Dataset}~\cite{perazzi2016benchmark} comprises high-resolution video sequences capturing a wide array of dynamic scenes and moving objects. It includes scenarios ranging from human activities and animal movements to complex object interactions and natural phenomena, providing a wide variety of motion patterns and object appearances. DAVIS doesn't contain depth information. We leverage \textit{Depth Anything V2} \cite{depth_anything_v2} to estimate depth, ensuring compatibility with our pipeline, and select 6 dynamic sequences for evaluation.
\end{itemize}

\subsubsection{Evaluation Metrics}

For camera tracking performance, we use the Root Mean Square Error (RMSE) and Standard Deviation (S.D.) of the Absolute Trajectory Error (ATE). 
For the quantitative evaluation of 3D reconstruction, we use accuracy (cm), completeness (cm), and completion rate (\%) thresholds set to $5$ cm.
In addition, we employ image rendering quality metrics, including Peak Signal-to-Noise Ratio (PSNR), Structure Similarity Index Measure (SSIM), and Learned Perceptual Image Patch Similarity (LPIPS). Since the original images contain both static backgrounds and dynamic foregrounds, the rendering metrics can effectively reflect the performance of dynamic-static composition mapping. 

\subsubsection{Baselines}
We conduct extensive and comprehensive comparisons between CAD-SLAM and traditional SLAM methods, NeRF-based SLAM methods, and 3D Gaussian-based SLAM methods to highlight the superiority of our method.


\subsubsection{Implementation Details}
All experiments are conducted on a server equipped with an Intel Platinum 8362 CPU and an NVIDIA A100 GPU.

\noindent \textbf{Tracking settings.} 
We employ a weighted combination of color and depth losses for tracking, with weights set to $\lambda_{track} = 0.6$ and $\lambda_{ssim} = 0.2$. Each frame is optimized for 100 iterations. The learning rates for camera pose optimization are set to 0.002 for rotation and 0.01 for translation. To mitigate the influence of outliers, alpha and depth filtering are applied to generate masks during the computation of the tracking loss. 

\noindent \textbf{Mapping settings.} 
For mapping, the order of spherical harmonics $L$ is 0. The position learning rate decays from an initial value of 0.001 to a final value of $1.6 \times 10^{-6}$. The learning rates for color, opacity, scaling, and rotation are set to 0.0025, 0.05, 0.005, and 0.001, respectively. Similar to tracking, each frame undergoes 100 iterations. The opacity threshold is set to 0.8 during densification and 0.3 during pruning. The weights for the composite losses are configured as $\lambda_{\text{ssim}} = 0.2$, $\lambda_{\text{color}} = 1.0$, $\lambda_{\text{depth}} = 1.0$, and $\lambda_{\text{reg}} = 1.0$. 

\noindent \textbf{Adaptive dynamic detection and tracking settings.} 
The thresholds for color and geometric inconsistencies $\tau_I = 20\cdot \operatorname{median}(I_{err})$ and $\tau_D =20\cdot \operatorname{median}(D_{err})$. The opacity thresholds for tracking $\tau_{track}$ and mapping $\tau_{map}$ are set to 0.7 and 0.8, respectively.
Adaptive dynamic object detection is performed every 5 frames for the Bonn, Wild-SLAM, and DAVIS datasets, and every 10 frames for the TUM RGB-D dataset. For 2D dynamic tracking, tracking is terminated if the center of a dynamic object approaches within 4\% of the image boundary. Additionally, tracking is considered erroneous and is terminated if the dynamic object’s mask area increases by more than 1.5× or its center moves over 20\% of the field of view within a single frame.

\subsection{Experimental Results}
\label{sec:exp_res}

\subsubsection{Camera Tracking Evaluation}
Tab.~\ref{tab_bonn_tracking}, Tab.~\ref{tab_tum_tracking}, and Tab.~\ref{tab_wild_tracking} present the camera tracking results of different methods on the Bonn, TUM RGB-D, and Wild-SLAM datasets, respectively. Results for other methods are taken directly from their original publications or obtained by running their official open-source code. 
Our method achieves the best tracking accuracy in most scenarios, with overall performance surpassing the current state-of-the-art methods. This is attributed to the precise masking of dynamic regions, which effectively avoids pose drift caused by motion interference. Compared to the semantic-prior-based methods DG-SLAM and DynaMoN, our method reduces tracking errors by 49.2\% and 41.9\%, respectively, demonstrating its advantage in adaptively identifying dynamic objects without relying on pre-defined dynamic categories.

\begin{figure*}[t] 
\center{\includegraphics[width=1.0\textwidth]{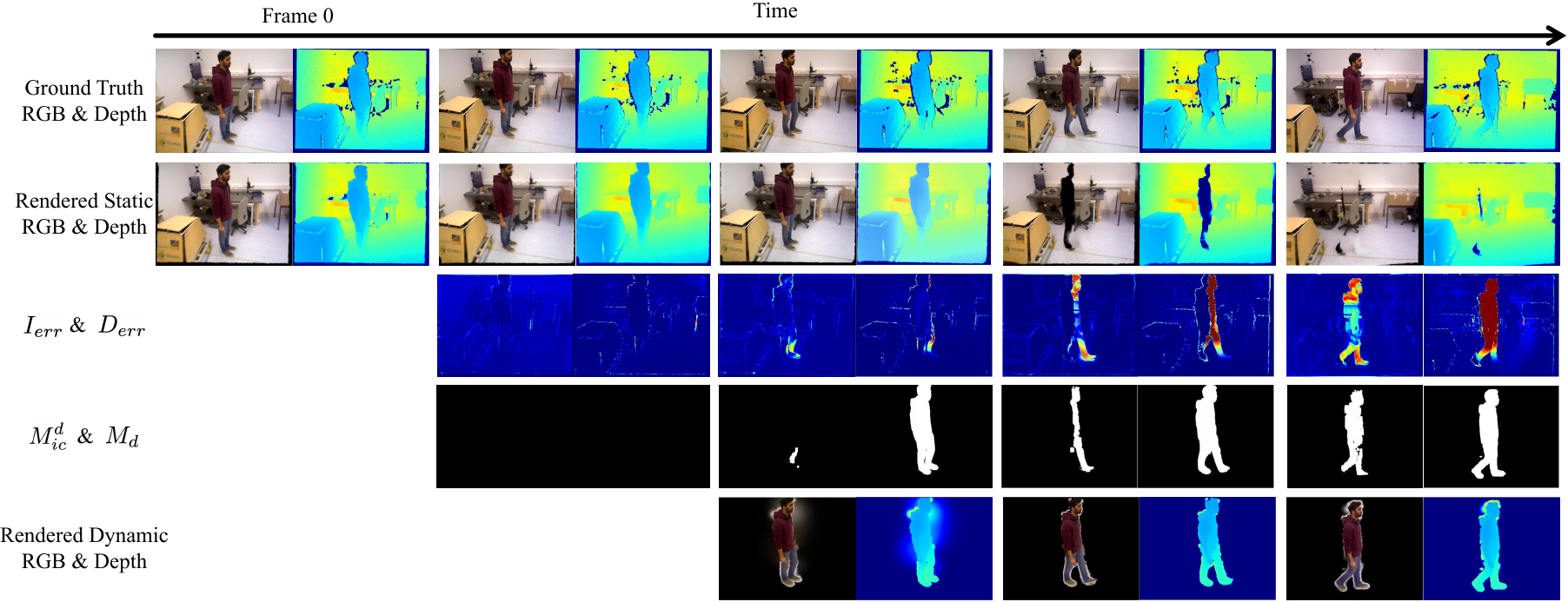}} 
\caption{Visualizations of the adaptive dynamic segmentation and dynamic-static separation mapping process during CAD-SLAM execution on Scene person tracking of the Bonn dataset. The first
frame is initialized as a static Gaussian map. Subsequently, adaptive dynamic segmentation
is performed based on inconsistency detection. When a dynamic
object is detected, it is separated from the static map, and a
sequential dynamic Gaussian model is constructed for dynamic
mapping. As the object moves, holes in the static map are
progressively filled in.}
\label{fig_bonn_dynamic}
\end{figure*}

\begin{table*}[t]
\centering
\caption{Quantitative comparison of rendering performance on bonn and tum dataset. The best results are denoted in \textbf{bold} font.}
\label{tab_bonn_tum_render}
\resizebox{0.9\textwidth}{!}{
\begin{tabular}{llccccccc}
\hline
\multirow{2}{*}{Methods} & \multirow{2}{*}{Metrics} & \multicolumn{3}{c}{Bonn} & \multicolumn{3}{c}{TUM RGBD}
\\
& & balloon & person\_track & person\_track2 & fr3/walk\_xyz  &fr3/walk\_static & fr3/sit\_hf & Avg. \\
\hline
\multirow{3}{*}{SplaTAM~\cite{keetha2024splatam}} 
& PSNR[dB] $\uparrow$ &  16.92 & 17.11\ & 15.54 & 17.15 & 18.54 &18.44 &17.28\\
& SSIM $\uparrow$ &  0.78 & 0.62 & 0.58 &0.66 &0.75& 0.75 & 0.69\\
& LPIPS $\downarrow$ &  \textbf{0.22}& 0.33 & 0.37 & 0.35 & 0.26 & 0.25 &0.30\\
\hline
\multirow{3}{*}{MonoGS~\cite{Matsuki:Murai:etal:CVPR2024}} 
& PSNR[dB] $\uparrow$ & 20.25 & 19.64 &18.47 & 12.82 & 16.35 &19.61 &17.86\\
& SSIM $\uparrow$ & 0.77 & 0.76 & 0.71 & 0.38 & 0.66 &0.72 &0.67\\
& LPIPS $\downarrow$ &0.34& 0.37 & 0.40 &0.55 & 0.27 &0.29&0.37\\
\hline
\multirow{3}{*}{CAD-SLAM(Ours)} 
& PSNR[dB] $\uparrow$ & \textbf{22.91} &\textbf{23.53} & \textbf{22.95} & \textbf{23.03} & \textbf{27.51} & \textbf{25.58} &\textbf{24.25}\\
& SSIM $\uparrow$ & \textbf{0.91}&\textbf{0.91}& \textbf{0.91} &\textbf{0.89} &\textbf{0.97} & \textbf{0.93}&\textbf{0.92}\\
& LPIPS $\downarrow$ &  \textbf{0.22}  &\textbf{0.25} &\textbf{0.24}&\textbf{0.24}&\textbf{0.09 }&\textbf{0.19} & \textbf{0.21}\\
\hline
\end{tabular}}
\end{table*}

\begin{table}
  \centering
     \caption{Comparison of different methods in terms of running time (ms).}
    \centering
    \resizebox{0.5\textwidth}{!}{
    \begin{tabular}{lccc}
        \toprule
        Methods & Dynamic Seg. & Tracking  & Mapping \\
        \midrule
        Rodyn-SLAM\cite{jiang2024rodyn} & 278.66 & 875.70 & 1083.60 \\
        SplaTAM\cite{keetha2024splatam} & - & 2630.36 & 548.06  \\
        WildGS-SLAM\cite{zheng2025wildgs} & 25.77 & 2467.28 & 2948.27 \\
        CAD-SLAM(Ours) & 68.79 & 1025.39 & 1108.78  \\
        \bottomrule
    \end{tabular}}
    \label{time}
  
\end{table}%

\subsubsection{3D Reconstruction Evaluation}
Most existing methods focus primarily on static scene reconstruction and neglect dynamic objects, whereas our method achieves composite reconstruction of both dynamic and static components. For a fair comparison, we evaluate only the static map. We employ Accuracy, Completeness, and Completion Ratio to assess geometric reconstruction quality, as shown in Tab.~\ref{tab_bonn_3d}. In the dynamic sequences of the Bonn dataset, CAD-SLAM achieves an average reconstruction accuracy of 6.30 cm, which is 21.8\% higher than that of DG-SLAM (8.06 cm), and a completeness of 7.64 cm, representing a 50.6\% improvement over DG-SLAM (15.46 cm). These gains stem from our method's more accurate adaptive dynamic identification and the dynamic-static separation mapping strategy, which prevents interference from missed dynamic objects and avoids missing static map areas due to false detections.

\subsubsection{Dynamic Separation Evaluation}
Fig.~\ref{fig_bonn_dynamic} visualizes the adaptive dynamic segmentation and dynamic-static separation mapping process during the execution of CAD-SLAM. The first frame is initialized as a static Gaussian map. It should be noted that our method does not enforce the first frame scene to be completely static. Subsequently, adaptive dynamic segmentation is performed based on inconsistency detection. When a dynamic object is detected, it is separated from the static map, and a sequential dynamic Gaussian model is constructed for dynamic mapping. As the object moves, holes in the static map are progressively filled in.

Fig.~\ref{fig_bonn_mask_dg} compares CAD-SLAM's masks with those of DG-SLAM~\cite{xu2024dg}. DG-SLAM~\cite{xu2024dg} employs semantic segmentation priors combined with a multi-view depth-warping mask to compensate for missing objects. Even objects that are static according to the prior are incorrectly segmented as dynamic regions. Although the multi-view depth-warping mask can compensate for missing object information, occlusions caused by viewpoint changes are also included in this mask and cannot be distinguished from truly dynamic areas, as indicated by the red circles in Fig.~\ref{fig_bonn_mask_dg}.

Fig.~\ref{fig_bonn_mask_wild} compares our dynamic masks with the uncertainty of WildGS-SLAM~\cite{zheng2025wildgs} on the Bonn dataset. As highlighted by the red circles, WildGS-SLAM exhibits significant under-detection and false detection. Blurred and incomplete uncertainty boundaries can lead to residual artifacts from dynamic objects persisting in the static map, while incorrectly overestimated uncertainty results in insufficient optimization of the corresponding static map regions. In contrast, our method accurately segments dynamic objects. Fig.~\ref{fig_davis_render} compares our dynamic masks with the uncertainty of WildGS-SLAM~\cite{zheng2025wildgs} across richer scenes on the DAVIS dataset. The results show that our dynamic masks are more complete and precise than the uncertainty masks of WildGS-SLAM~\cite{zheng2025wildgs} and remain unaffected by background interference. The successful segmentation of various objects across multiple environments robustly demonstrates the adaptability of our method.

\subsubsection{Rendering Quality Evaluation}
We employ PSNR, SSIM, and LPIPS to quantitatively evaluate rendering quality, with results presented in Tab.~\ref{tab_bonn_tum_render}. Our method achieves state-of-the-art rendering performance, benefiting from the precise modeling of foreground details through dynamic-static decoupled mapping and the artifact-free reconstruction of the static background. Fig.~\ref{fig_bonn_render} provides a visual comparison of renderings from different methods. SplaTAM~\cite{keetha2024splatam} and MonoGS\cite{Matsuki:Murai:etal:CVPR2024} exhibit severe blurring and geometric distortion. WildGS-SLAM~\cite{zheng2025wildgs} can only render the static scene, and residual artifacts remain due to ambiguity at dynamic boundaries. In contrast, our CAD-SLAM accurately reconstructs texture details of the static background while preserving the complete form of dynamic objects (e.g., balloon contours, human posture), achieving high-quality rendering that seamlessly integrates static and dynamic elements. Fig.~\ref{fig_davis_render} compares our renderings with those of the current state-of-the-art method, WildGS-SLAM~\cite{zheng2025wildgs}, across diverse scenarios. As shown, due to the failure of its dynamic uncertainty prediction, WildGS-SLAM cannot effectively distinguish dynamic objects, resulting in blurry dynamic regions, such as the camel area in the "camel" sequence and the car area in the "car-turn" sequence. In comparison, our method can accurately identify and model various categories of dynamic objects, demonstrating superior adaptability.

\begin{figure*}[t] 
\center{\includegraphics[width=1.0\textwidth]{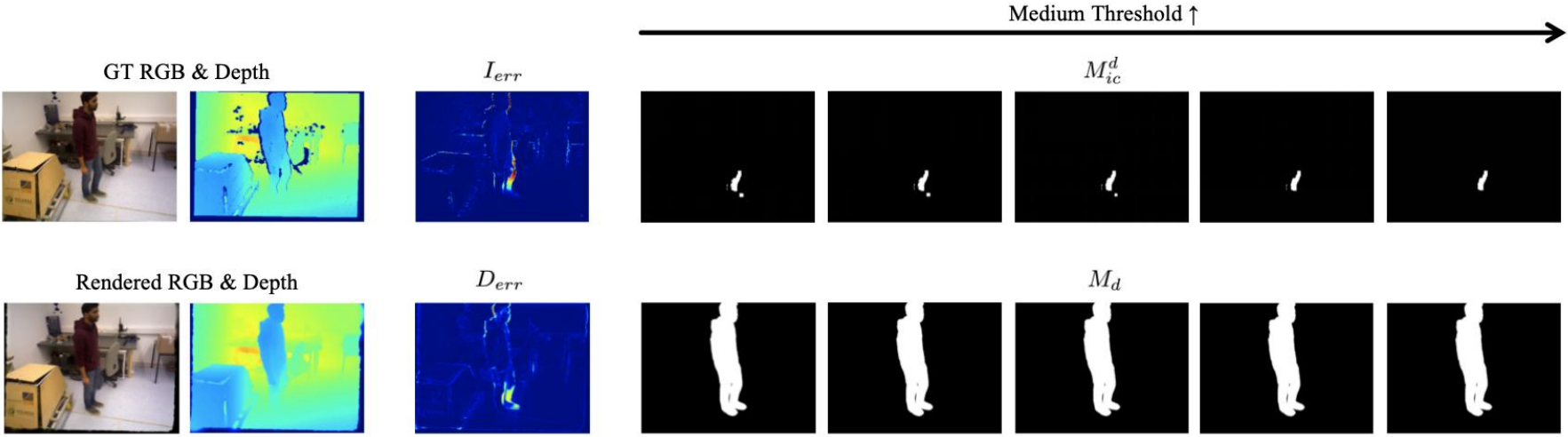}} 
\caption{Inconsistent regions $M^d_{ic}$ and dynamic masks $M_d$ under different multiples of the error median. From left to right: 10×, 15×, 20×, 25×, and 30× the error median. The
dynamic mask segmentation results remain unaffected, with
true motion regions consistently identifiable. This demonstrates the effectiveness and robustness of our method.}
\label{fig_abl_incon}
\end{figure*}

\begin{table}
  \centering
\caption{Comparison of tracking metrics of different ways to manage dynamic segmentation. }
    \resizebox{1.0\linewidth}{!}{
    \begin{tabular}{lcccccc}
    \toprule
    Methods & \multicolumn{2}{c}{\texttt{balloon}} & \multicolumn{2}{c}{\texttt{ball\_track}} & \multicolumn{2}{c}{\texttt{mv\_box2}}  \\
    \midrule
     & \textit{RMSE} & \textit{SD} & \textit{RMSE} & \textit{SD} & \textit{RMSE} & \textit{SD} \\
         a. w/o dynamic seg. & 52.5 & 21.6 & 15.2 & 5.2 & 18.3 & 5.4\\
    b. w/ MaskDINO\cite{li2022mask} & 5.5 & 2.1 & 11.4 & 4.6 & 12.5 & 3.9  \\

    c. w/o keyframe DBA & 3.3 & \textbf{1.0} & 6.9 & 3.9 & 7.4 & 3.0 \\
    d. CAD-SLAM (Ours) & \textbf{2.7} & \textbf{1.0} & \textbf{3.4} &\textbf{ 1.2} &\textbf{ 2.1} &\textbf{ 0.8 }  \\
    \bottomrule
    \end{tabular}
    }
    \label{ablation}
\end{table}%

\subsubsection{Runtime Analysis.}
Tab. \ref{time} presents the runtime analysis, including the time for dynamic segmentation, camera tracking, and mapping. Our tracking and mapping times are comparable to other methods, while additionally performing dynamic mapping. For dynamic segmentation, our method is more efficient than Rodyn-SLAM\cite{jiang2024rodyn}, which relies on pre-trained semantic segmentation and optical flow networks.

\begin{figure}
\center{\includegraphics[width=0.45\textwidth]{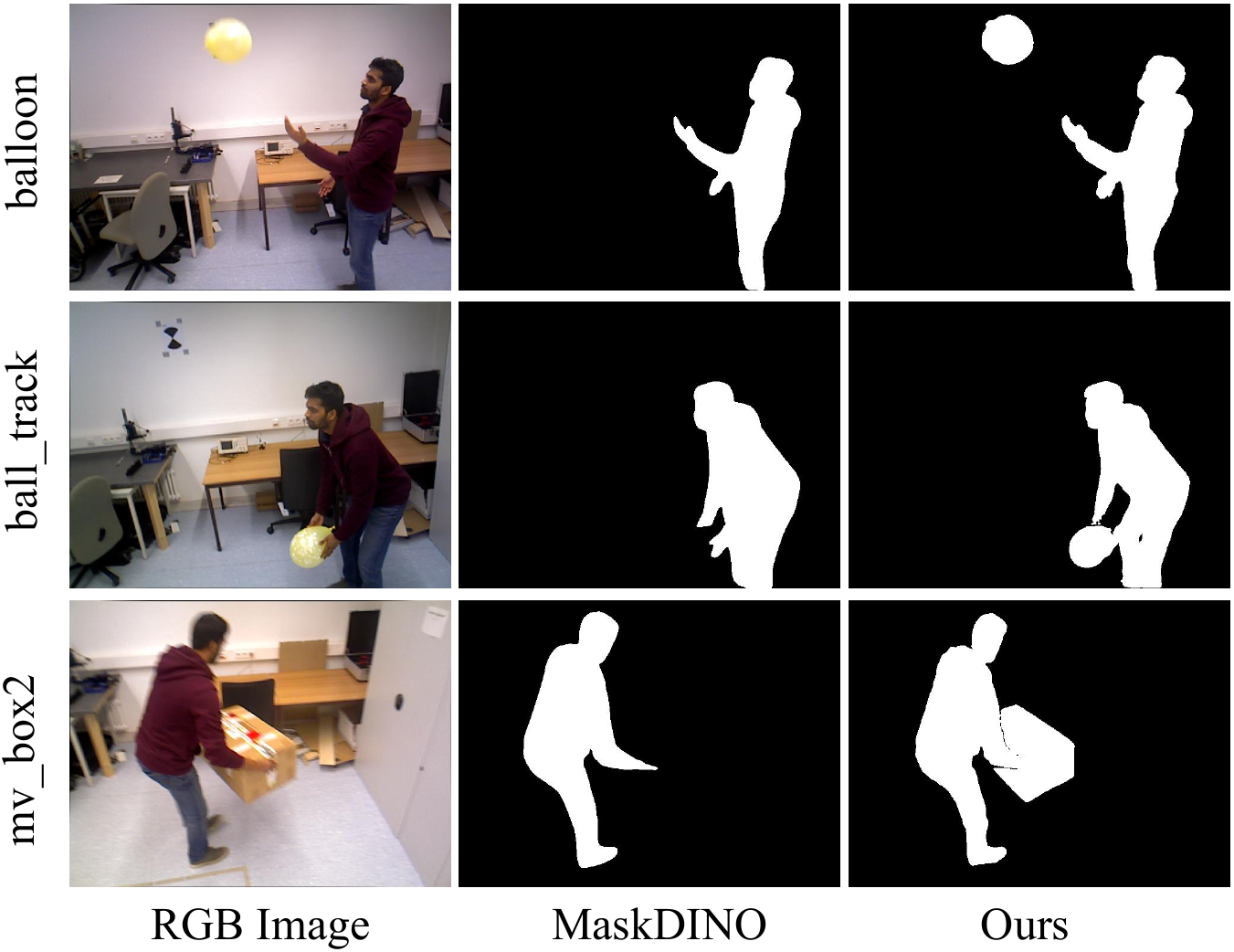}} 
\caption{Comparison of the dynamic segmentation results of our method with those obtained using a semantic segmentation network. Unlike prior-based semantic segmentation, our method can adaptively detect atypical moving objects, such as boxes and balloons, enabling more precise and flexible dynamic segmentation.}
\label{fig_abl_mask}
\end{figure}

\subsection{Ablation Study}

\begin{figure}[h] 
\center{\includegraphics[width=1.0\linewidth]{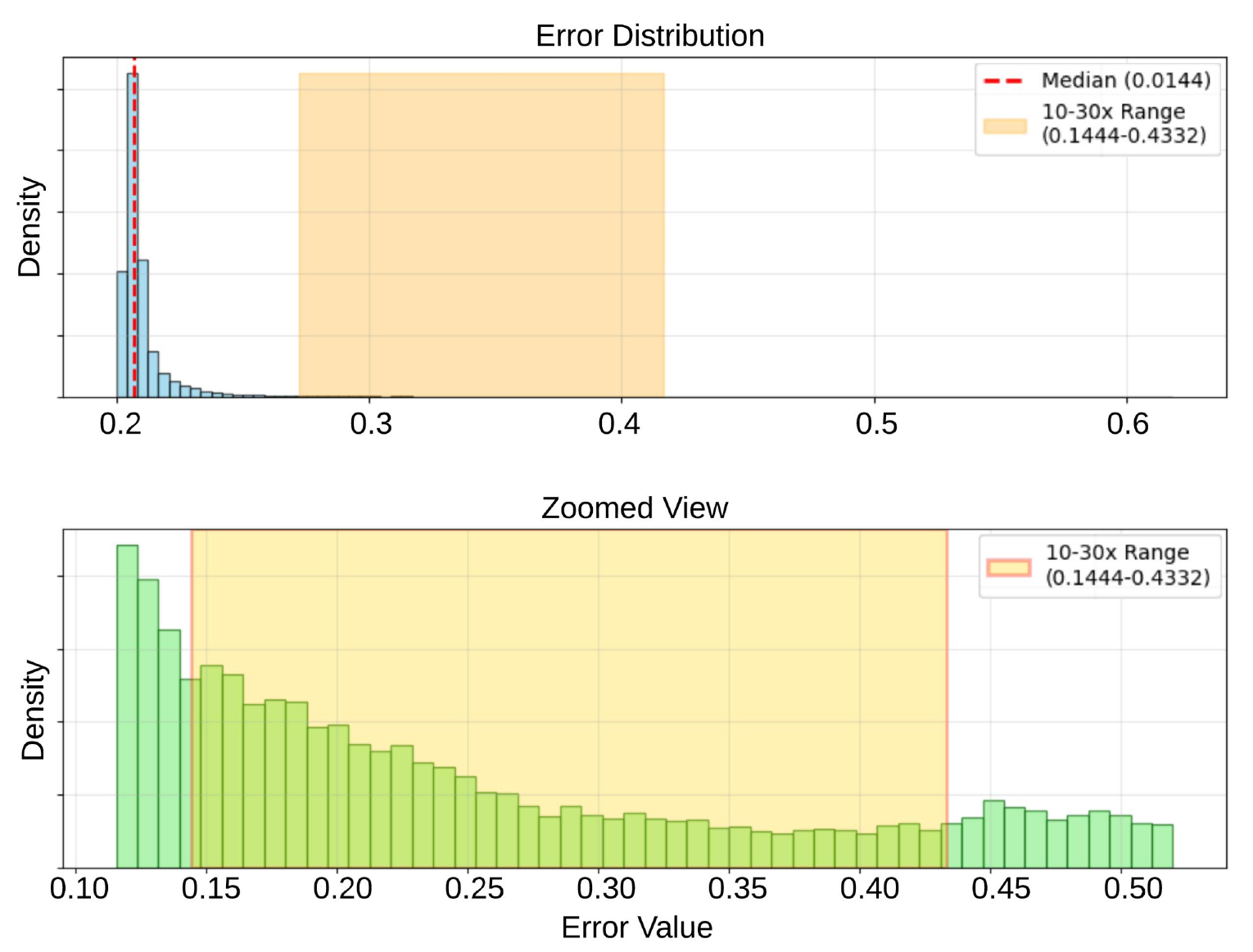}} 
\caption{Rendered error distribution analysis. The red dashed lines indicate error medians, while the yellow shaded areas represent the range from 10× to 30× the error median. 
Selecting 20× the median as the operational threshold effectively distinguishes genuine motion areas. Notably, the error threshold maintains a wide acceptable range (10×-30×, even wider), ensuring robust applicability in practice. }
\label{fig_render_error}
\end{figure}
We conduct an ablation study on our adaptive dynamic segmentation, to demonstrate whether an explicit segmentation is necessary.
Fig.~\ref{fig_abl_mask} presents a comparison between our method's dynamic segmentation results and those obtained by a semantic segmentation network. We utilize MaskDINO\cite{li2022mask}, with the predefined dynamic category set to ``human". Unlike prior-based semantic segmentation, our method can adaptively detect atypical moving objects, such as boxes and balloons, enabling more precise and flexible dynamic segmentation.
Tab.~\ref{ablation} shows the camera tracking results under different settings: a. without dynamic segmentation, b. using the prior-based semantic segmentation, and d. employing our adaptive dynamic segmentation. With more accurate and flexible dynamic segmentation, camera tracking accuracy is significantly improved. Additionally, we validated the effect of keyframe DBA. As shown in row c. of Tab.~\ref{ablation}, keyframe DBA can significantly improve tracking accuracy.

Additionally, we conducted a parameter sensitivity analysis on the adaptive dynamic segmentation module. For geometric and color inconsistency thresholds, we statistically analyze the rendered geometric and color error distributions, as shown in Fig.~\ref{fig_render_error}. The red dashed lines indicate error medians, while the yellow shaded areas represent the range from 10× to 30× the error median. The observed long-tailed error distribution reveals two distinct characteristics: minimal errors in static regions (distribution head) and significantly larger errors in dynamic regions (distribution tail). Selecting 20× the median as the operational threshold effectively distinguishes genuine motion areas. Notably, the error threshold maintains a wide acceptable range (10×-30×, even wider), ensuring robust applicability in practice. Fig.~\ref{fig_abl_incon} presents a sensitivity analysis of the threshold selection, demonstrating that while the inconsistent mask area slightly decreases from 10× to 30× median thresholds. The dynamic mask segmentation results remain unaffected, with true motion regions consistently identifiable. 
This analysis demonstrates the effectiveness and robustness of our threshold Settings. 

\section{Conclusion}
\label{sec:conclusion}

We propose CAD-SLAM, a novel dynamic dense visual SLAM system that addresses the core challenges of dynamic environment perception and modeling. By introducing a consistency-aware dynamic detection mechanism, we eliminate reliance on brittle semantic priors and achieve category-agnostic, real-time dynamic object identification. Complemented by a bidirectional tracking strategy and dynamic-static decoupled mapping framework, our method not only maintains state-of-the-art performance in camera localization and static scene reconstruction but also enables accurate modeling of dynamic objects. Extensive experiments across four real-world datasets validate the robustness, adaptability, and efficiency of CAD-SLAM, providing a reliable technical foundation for robotic perception and interaction in complex dynamic scenarios.
Despite these advances, several promising directions remain for further refinement. First, develop a unified end-to-end network that integrates scene consistency analysis and dynamic segmentation, eliminating reliance on external models like MobileSAM. Second, integrate motion prediction models into the tracking pipeline. By predicting the future motion of dynamic objects, the system can handle temporary occlusion and provide more support for downstream tasks. Additionally, the real-time performance could be further improved.

\bibliographystyle{IEEEtran}  
\bibliography{IEEEabrv,bare_jrnl} 

\vfill

\end{document}